%% file: main.tex
\documentclass[runningheads]{llncs}

\usepackage{eccv}

\usepackage{eccvabbrv}

\usepackage{graphicx}
\usepackage{booktabs}

\usepackage[accsupp]{axessibility}  % Improves PDF readability for those with disabilities.
\input{preamble}

\usepackage{hyperref}

\usepackage{orcidlink}

\begin{document}

% ---------------------------------------------------------------

\title{An accurate detection is not all you need \\ to combat label noise in web-noisy datasets}
\titlerunning{PLS-LSA}
\author{Paul Albert\inst{1}, Jack Valmadre\inst{1}, Eric Arazo\inst{2}, Tarun Krishna\inst{3},\\Noel E. O'Connor\inst{3}, Kevin McGuinness\inst{3}\\}
\institute{Australian Institute for Machine Learning, University of Adelaide \and
CeADAR: Ireland’s Centre for Applied Artificial Intelligence \and
Insight Centre for Data Analytics, Dublin City University \\ \email{paul.albert@adelaide.edu.au}}

%\author{First Author\inst{1}\orcidlink{0000-1111-2222-3333} \and
%Second Author\inst{2,3}\orcidlink{1111-2222-3333-4444} \and
%Third Author\inst{3}\orcidlink{2222--3333-4444-5555}}

\authorrunning{Albert, P. et al.}

%\institute{Princeton University, Princeton NJ 08544, USA \and
%Springer Heidelberg, Tiergartenstr.~17, 69121 Heidelberg, Germany

\maketitle
\input{sec/0_abstract}    
\input{sec/1_intro}
\input{sec/2_related}
\input{sec/3_lsa}
\input{sec/4_exp}
\input{sec/5_conclu}

\input{sec/X_suppl}
\clearpage
\bibliographystyle{splncs04}
\bibliography{main}


\end{document}

%% file: preamble.tex
\usepackage[dvipsnames]{xcolor}
\usepackage{graphicx}
\usepackage{amsmath}
\usepackage{amssymb}
\usepackage{booktabs}
\usepackage{multirow}
\usepackage{comment}
\usepackage{xcolor}
\usepackage{wrapfig}

\usepackage{pifont}
\newcommand{\cmark}{\ding{51}}%
\newcommand{\xmark}{\ding{55}}%
\newcommand{\bx}{\mathbf{x}}
\newcommand{\by}{\mathbf{y}}
\newcommand{\bz}{z}
\newcommand{\bp}{p}

\usepackage{tabularx}

%% file: sec/0_abstract.tex
\begin{abstract}
    Training a classifier on web-crawled data demands learning algorithms that are robust to annotation errors and irrelevant examples.
   This paper builds upon the recent empirical observation that applying unsupervised contrastive learning to noisy, web-crawled datasets yields a feature representation under which the in-distribution (ID) and out-of-distribution (OOD) samples are linearly separable \cite{2022_ECCV_SNCF}.
   We show that direct estimation of the separating hyperplane can indeed offer an accurate detection of OOD samples, and yet, surprisingly, this detection does not translate into gains in classification accuracy.
   Digging deeper into this phenomenon, we discover that the near-perfect detection misses a type of clean examples that are valuable for supervised learning.
   These examples often represent visually simple images, which are relatively easy to identify as clean examples using standard loss- or distance-based methods despite being poorly separated from the OOD distribution using unsupervised learning.
   Because we further observe a low correlation with SOTA metrics, this urges us to propose a hybrid solution that alternates between noise detection using linear separation and a state-of-the-art (SOTA) small-loss approach.
   % Our clean-sample selection strategy can serve as a practical replacement to label noise approaches training a noise-robust contrastive objective.
   When combined with the SOTA algorithm PLS, we substantially improve SOTA results for real-world image classification in the presence of web noise
   \url{https://github.com/PaulAlbert31/LSA}
\end{abstract}

%% file: sec/1_intro.tex
\section{Introduction}
\label{sec:intro}
Developing learning algorithms that are robust to label noise promises to enable the use of deep learning for a variety of tasks where automatic but imperfect annotation is available. This paper studies the specific case of web-noisy datasets for image classification. A web-noisy dataset~\cite{2017_arXiv_WebVision,2021_ICCV_weblyfinegrained} is in fact the starting point for most generic image classification datasets, before human curation and label correction is conducted. To create a web-noisy dataset, the only required human intervention is the definition of a set of classes to be learned. Once the classes are defined, examples are recovered by text-to-image search engines, sometimes aided by query expansion and image-to-image search. Since the text surrounding an image on a web-page may not be an accurate description of the semantic content of the image, some training examples will incorrectly represent the target class, leading to a degradation of both the model's internal representation and its final decision.
Research has identified that, in the case of web-noisy datasets, out-of-distribution (OOD) images are by far the most dominant form of noise~\cite{2022_WACV_DSOS}.

We propose to build upon the observations made in SNCF~\cite{2022_ECCV_SNCF}, who observed that representations learned by unsupervised contrastive algorithms on OOD-noisy datasets displayed linear separability between in-distribution (ID) and OOD images. We extend this observation to web-noisy datasets containing OOD images where we notice that the separation is not as good as SNCF observed on synthetically corrupted datasets. Upon further investigation, we however notice that the separation is recovered when evaluating intermediate representations, computed earlier in the network.
Another limitation of SNCF we aim to address is the reliance on clustering to retrieve the noisy samples so we propose to directly estimate the linear separator. We compute an approximated linear separation using SOTA noise-robust algorithms~\cite{2021_ICCV_RRL,2023_WACV_PLS} to obtain an imperfect clean/noisy detection, which we then use to train a logistic regression on the unsupervised contrastive features. This produces an accurate web-noise detection.

Interestingly, when substituting our more accurate noise detection for the original detection metric in naive ignore-the-noise algorithms and subsequently the noise robust algorithm PLS~\cite{2023_WACV_PLS}, we observe a decrease in classification accuracy. In fact, we identify that few simple yet important clean examples are missed by our linear separation although they are correctly retrieved by SOTA noise detection~\cite{2023_WACV_PLS,2021_ICCV_RRL}. Because we find that our linear separation is decorrelated these SOTA noise detectors, we propose a detection strategy that combines linear separation (which achieves high specificity and sensitivity) and SOTA noise-detection approaches (which correctly retreive those few important samples) by alternating each every epochs. We combine this noise detection with PLS to create PLS-LSA which we find to be superior to existing noise-robust algorithms on a variety of classification tasks in the presence of web-noise. We contribute:
\begin{enumerate}    
    \item A novel noise detection approach that extends the work of SNCF~\cite{2022_ECCV_SNCF} to web-noisy datasets where we improve the detection of OOD samples present in web-noise datasets by explicitly estimating the linear separation between ID and OOD samples. We demonstrate that this detection strategy is weakly correlated to existing small-loss and distance-based approaches.
    \item An investigation into the disparity between noise retrieval performance and classification accuracy of noise-robust algorithms.
    \item A novel noise correction approach, Linear Separation Alternating (LSA), that combines linear separation with uncorrelated SOTA noise detection.
    \item A series of experiments and ablation studies, including a voting co-training strategy PLS-LSA+ that concurrently trains two models. We conduct these experiments on controlled and real-world web-noisy datasets to demonstrate the efficacy of our algorithm PLS-LSA.
\end{enumerate}

%% file: sec/2_related.tex
\section{Related work}
\iffalse
\subsection{Noise robust regularization} 
A first approach to noise robust training is to design noise-robust losses. The Gold and Forward losses~\cite{2017_CVPR_ForwardLoss,2018_NIPS_GoldLoss} enforce a noise prior at the class level by multiplying the classifier's predictions with an estimated noise matrix. Early Learning Regularization (ELR)~\cite{2020_NeurIPS_EarlyReg} regularizes a cross-entropy classification loss by penalizing high dot product between the prediction of the network and its potentially noisy target. The gradient of the combined cross-entropy and dot product regularization is shown to reduce over-fitting of training samples that the network disagreed with in early training steps. The Taylor Cross-Entropy (TCE) loss~\cite{2021_IJCAI_TCE} decomposes the categorical cross-entropy loss using the Taylor series, and it was observed that label-noise robustness can be achieved at particular decomposition orders. Finally, mixup~\cite{2018_ICLR_Mixup} is a noise-robust data augmentation that linearly interpolates pairs of training images and their classification label, and was shown to limit the impact of label noise on classification accuracy~\cite{2019_ICML_BynamicBootstrapping}. An interesting aspect of naturally robust losses is that they are easily complementary to orthogonal approaches to combat label noise and mixup in particular is systematically used.
\fi
\subsubsection{Detection and correction of incorrect labels}

The most popular approach to tackle label noise is to explicitly detect ID samples with incorrect labels, either because they are harder to learn than their clean counterparts or because they are distant from same-class training samples in the feature space. Noise detection strategies include evaluating the training loss~\cite{2019_ICML_BynamicBootstrapping,2020_ICLR_DivideMix,2021_arXiv_propmix,2021_ICPR_towardsrobust}, the Kullback-Leibler or Jensen-Shannon divergence between prediction and label~\cite{2021_CVPR_JoSRC}, the entropy or the confidence of the prediction~\cite{2022_WACV_DSOS,2020_ICML_MentorMix} or the consistency of the prediction across epochs~\cite{2019_ICLR_Forgetting}. An alternative is to measure the distance between noisy and clean samples in the feature space: RRL~\cite{2021_ICCV_RRL} detects noisy samples as having many neighbors from different classes and NCR~\cite{2022_CVPR_NeighborConsistency} regularizes training samples with similar feature representations to have similar predictions, reducing noisy label overfitting. We also note here that all recent label noise algorithms utilize the mixup~\cite{2018_ICLR_Mixup} regularization which has proven to be highly robust to label noise~\cite{2019_ICML_BynamicBootstrapping}. 

While many noise-robust algorithms have proposed loss-based or distance-based noise detection metrics, the distinct advantages and biases of each strategy remain unexplored. Furthermore, considering that loss-based and distance-based detections are sufficiently decorrelated, combining the strengths of these distinct metrics is appealing, yet has not been previously explored. This paper observes the decorrelation of some noise detection metrics and proposes a non-trivial combination that improves generalization noise-robust algorithms over either metric taken independently.
\vspace{-10pt}
\subsubsection{Out-of-distribution noise in web-noisy datasets}
In web-noisy datasets, OOD (or \emph{open-world}) noise is the dominant type~\cite{2022_WACV_DSOS}. Since ID noise is still present in small amounts in web-noisy datasets, algorithms propose concurrently detect ID and OOD noise. EvidentialMix~\cite{2020_WACV_EDM} and DSOS~\cite{2022_WACV_DSOS} use specialised losses that exhibit three modes when evaluated over all training samples. Each of the modes are observed to mostly contain clean, OOD and ID noisy samples. A mixture of gaussians is then used to retrieve each noise type. SNCF~\cite{2022_ECCV_SNCF} observed that unsupervised contrastive learning trained on a web-noisy dataset learns representations that are linearly separated between ID and OOD samples and use a clustering strategy based on OPTICS~\cite{1999_ACM_Optics} to retrieve each noise type. The linear separability of in-distribution (ID) and out-of-distribution (OOD) representations noted in SNCF holds promise but has yet to be transitioned from synthetically corrupted to web-noisy data. This paper aims to address this gap.
\vspace{-10pt}
\subsubsection{Unsupervised learning and label noise}
Optimizing a noise robust (un)-supervised contrastive objective together with the classification loss can help improve the representation quality as well as detect OOD samples in the feature space. ScanMix learns SimCLR representations as a starting point for noise-robust contrastive clustering~\cite{2023_PR_scanmix} and SNCF~\cite{2022_ECCV_SNCF} observes linear separability on unsupervised iMix features~\cite{2021_ICLR_iMix}. Unsupervised contrastive features have also been used to initialize networks prior to noise-robust training in PropMix~\cite{2021_arXiv_propmix} and C2D~\cite{2022_WACV_C2D} or used as a regularization to the supervised objective in RRL~\cite{2021_ICCV_RRL}. While unsupervised initialization or regularization has been employed to enhance the generalization accuracy of networks trained under label noise, our primary focus lies in its ability to detect noisy samples before starting the supervised learning phase. We aim to enhance the linear separation observed in SNCF, particularly by extending it from synthetically corrupted to web-noisy datasets and by eliminating the requirement for clustering.

In this review, we find that unsupervised learning shows promise in identifying OOD images even before noise-robust supervised training begins. While many algorithms demonstrate an effective identification of OOD images in synthetically corrupted datasets, their generalization to web-noisy datasets is non-evident. Furthermore, although we observe a high correlation between loss-based and distance-based metrics, neither correlates directly with the linear detection observed in SNCF. We propose to evaluate the disparities between these metrics and to explore potential combinations to enhance noise detection, surpassing the capabilities of each metric taken independently.

\iffalse
\subsection{Contrastive learning and label noise}
In the case of noise robust supervised-contrastive learning, MOIT~\cite{2021_CVPR_MOIT} linearly combines correct and incorrect training samples in a mixup-inspired Interpolated Constrastive Learning (ICL) loss objective, and enforces that only the clean labels be learned. SNCF~\cite{2022_ECCV_SNCF} uses a supervised contrastive N-pairs objective~\cite{2021_ICLR_iMix} where similar OOD samples are encouraged to cluster together to learn low-level features, and PLS~\cite{2023_WACV_PLS} trains a smooth interpolation between a supervised and unsupervised contrastive objective where the noisiness of a sample encourages an unsupervised (OOD) or supervised (ID) behavior. Finally, RankMatch (RM)~\cite{2023_ICCV_rankmatch} trains a Rank Contrastive Loss that pulls together samples with similar activations in the contrastive feature space. 
\fi

%% file: sec/3_lsa.tex
\section{Linear Separation Alternating (LSA)}
This section details the contributions of this paper and the alternating noise detection strategy we use to combat label noise.
We consider in this paper the case of a noisy web-noisy image dataset $\mathcal{D} = \{\bx_i, \by_i\}_{i=1}^N$ of size $N$ where the images $\mathcal{X}=\{\bx_i\}_{i=1}^N$ are associated with a classification label $\{\by_i\}_{i=1}^N \in \{1, \dots C\}$. We denote vectors with bold letters. The classification labels are expected to be possibly mis-assigned, i.e. incorrectly characterize the target object in the image they are assigned to (label noise). The clean or noisy nature of the training samples is unknown. Our goal is to learn an accurate classifier $\Phi(\bx)$ that performs an accurate classification despite the label noise present in~$\mathcal{X}$. In our case, we consider that $\Phi$ is a neural network.

\subsection{Identifying OOD images in web-noisy datasets~\label{sec:separationsncf}}
This section proposes to detect OOD images in web-noisy datasets by building on the detection of SNCF~\cite{2022_ECCV_SNCF}. SNCF observes that an unsupervised algorithm trained on a web-noisy dataset containing OOD images will learn linearly separable representations for the ID and OOD samples in the dataset. While this is primarily an empirical observation, it was hypothesized to be a consequence of the uniformity and alignment principles of contrastive learning~\cite{2020_ICLR_understandingcontrastive}. The alignment principle in contrastive learning encourages samples with similar visual features to cluster together while the uniformity principle encourages training samples to be uniformly distributed in the feature space. OOD images cannot satisfy the alignment principle since they are visually different from all other images in the dataset and are pushed by all ID samples on one side on hypersphere, becoming linearly separable from the ID samples~\cite{2022_ECCV_SNCF}.
As an aside, this hypothesis implies that the linear separation may not occur when training on visually similar out-of-sample images, a problem which we will revisit in the following sections. Importantly, the separability of ID and OOD samples only occurs for samples the unsupervised algorithm is trained on and cannot generalize to new unseen OOD images. % In other words, the linear separability of OOD images does not generalize to the problem of OOD/anomaly detection.

\begin{figure}[t]
    \centering
    \includegraphics[width=.7\linewidth]{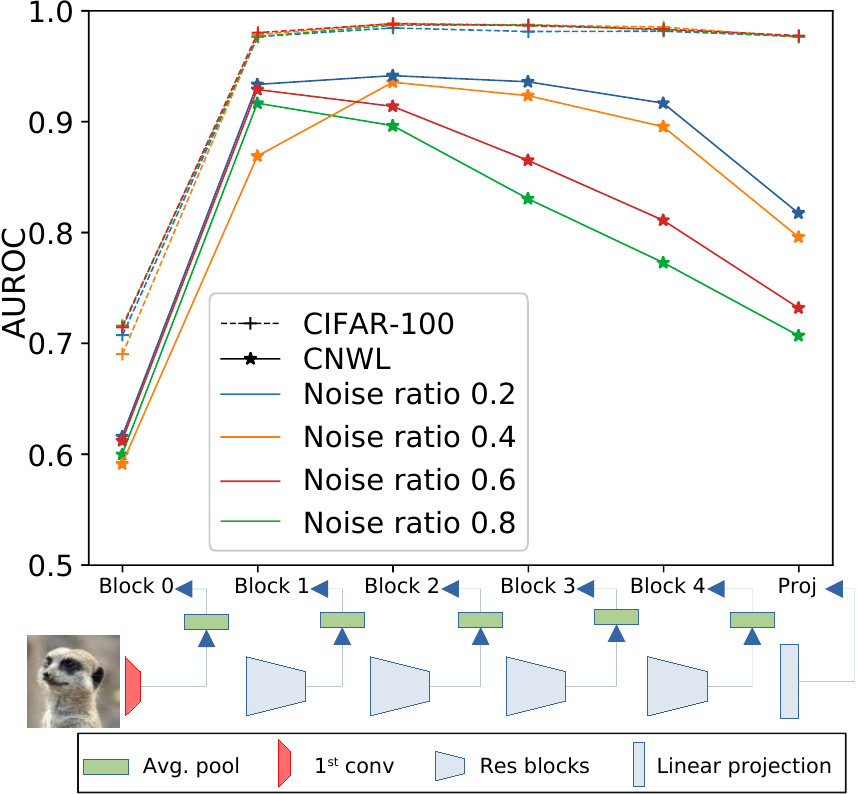}
    \caption{Extending the work of~\cite{2022_ECCV_SNCF} we observe that for web noise (CNWL), ID and OOD samples become more separable in earlier representations in the network\label{fig:linearsepdepth}}
\end{figure}

Although SNCF~\cite{2022_ECCV_SNCF} observed the ID/OOD separation on synthetically corrupted datasets, i.e CIFAR-100~\cite{2009_CIFAR} corrupted by ImageNet32~\cite{2017_arXiv_ImageNet32}, Figure~\ref{fig:linearsepdepth} further described below shows that we do not observe as good a separation when moving to the web-noisy CNWL~\cite{2020_ICML_MentorMix} but that the separability improves when looking at earlier representations. 
The weaker separability of OOD/ID images in web-noisy datasets compared to artificially corrupted datasets is explained by OOD images in web-noise datasets retaining weak semantic similarities with ID images. This is particularly true at the text level, which exhibited relevant similarities for the search engine during dataset creation. We propose that stronger separation occurs in low-level representations because they are more generic. Earlier representations easily align ID images of the same class due to shared low-level semantics, while OOD samples only become overfit in deeper layers, thus making separation increasly more difficult. A lesser corruption of earlier representations by label noise has for example been observed in~\cite{2021_ICPR_towardsrobust}.

\vspace{-15pt}
\subsubsection{Linear separation improves in deeper layers~\label{sec:sepdeeper}}
Our first contribution is then to observe that although the linear separation between ID and OOD is less evident in web-noisy datasets, it improves again when using earlier representations.
Figure~\ref{fig:linearsepdepth} gives an overview of the linear separability of ID and OOD images in the CNWL dataset~\cite{2020_ICML_MentorMix} (web-noise) compared to the CIFAR-100 dataset~\cite{2009_CIFAR} artificially corrupted with OOD images from ImageNet32~\cite{2017_arXiv_ImageNet32} using the unsupervised algorithm SimCLR~\cite{2020_ICML_SimCLR} to pre-train a PreActivation ResNet18~\cite{2016_ECCV_preactresnet}. To compute the lower level features, we average-pool then L2 normalize representations at the end of each ResNet block. To compute the linear separability, we utilize the clean-noisy oracle to train a non-penalized logistic regressor to predict noise from the unsupervised features. The linear regressor is then evaluated on a held-out noisy test set previously unseen by the regressor. We report the area under the ROC curve (AUROC) for the logistic regressor to identify correct/incorrect training samples. Our train/test split for evaluation of the linear classifier comprises $45,000$ training and $5,000$ testing images, and is constructed from the full $50,000$ training images available for the overall classification task, all of which were used in unsupervised representation learning.

\vspace{-15pt}
\subsubsection{Estimating the linear separator~\label{sec:estim}}
A straight-forward approach to estimating the linear separation is to task human annotators to label randomly selected samples as ID or OOD, thus fulfilling the oracle role in Section~\ref{sec:sepdeeper}. This strategy is usually referred to as learning to combat label noise with a trusted subset $\hat{\mathcal{T}} = \{\bx_i,\hat{\bz_i}\}_{i=1}^{K}$ where $\hat{\bz_i} = 1$ means that the image $\bx_i$ is OOD ($\hat{\bz_i} = 0$ for ID). Although we will show in the supplementary material that a good approximation for the linear separator can be achieved even given a small human-labeled subset of $100$ images, most state-of-the-art noise robust algorithms do not rely on ID/OOD human annotations. We thus propose to estimate $\hat{\mathcal{T}}$ in an unsupervised manner. The unsupervised strategy of SNCF~\cite{2022_ECCV_SNCF} is to use a clustering approach based on OPTICS~\cite{1999_ACM_Optics}. We propose in this paper to avoid clustering and instead to train the linear separation using an unsupervised ID/OOD subset $\hat{\mathcal{T}}$. 

We propose to build $\hat{\mathcal{T}}$ using unsupervised noise detection metrics $z(\bx_i, \by_i) = \hat{\bz_i}$~\cite{2020_ICLR_DivideMix, 2020_NeurIPS_EarlyReg, 2021_CVPR_MOIT, 2021_ICCV_RRL,2023_WACV_PLS}. We will examine recent examples of loss-based and distance-based noise detections later in the paper.
Given $\hat{\mathcal{T}}$ estimated from $Z=\{\bz_i\}_{i=1}^N$, we can train the linear regressor to effectively refine the estimated noisiness of a training sample $(\bx_i)$ to $\mathcal{L}_r(\bx_i, \bz_i) = w_i$ where $\mathcal{L}_{r}$ is a linear classification algorithm.
Effectively, given the initial noise detection $Z$ and the unsupervised contrastive features, we produce an improved one $W = \{w_i\}_{i=1}^N$, the linear-separation detection.

We find that, although an unsupervised $\hat{\mathcal{T}}$ contains detection errors, we still accurately estimate the linear separation due to the natural outlier robustness of linear classifiers. We additionally attempted to construct $\hat{\mathcal{T}}$ by selecting only the $M$ most confidently clean/incorrect samples according to the metric $z$ but found that it lead to a less accurate $W$.

\begin{table*}[t]
\centering{}\caption{Using multiple noise detection metrics to train a naive noise-ignoring algorithm on the CNWL datset. We report noise retreival performance and classification accuracy. None signifies training without noise removal. We bold the \textbf{best results} and underline \underline{the worst}, higher is better. Results averaged over 3 random seeds $\pm$ std\label{tab:strongercleannoisy}}
\global\long\def\arraystretch{0.9}%
\resizebox{0.9\linewidth}{!}{{{}}
\begin{tabular}{c>{\centering}c>{\centering}c>{\centering}c>{\centering}c>{\centering}c>{\centering}c>{\centering}c>{\centering}c}
\toprule
Noise ratio $\downarrow$ & Metric $\rightarrow$ & \textcolor{gray}{None} & PLS & RRL & $W_{PLS}$ & $W_{RRL}$ & \textcolor{gray}{Oracle} \tabularnewline
\midrule
\multirow{3}{*}{$0.2$} & AUROC & $-$ & $66.4  \scriptstyle \pm 0.2$ & $\underline{58.4  \scriptstyle \pm 0.2}$ & $\mathbf{84.9 \scriptstyle \pm 0.2}$ & $\mathbf{84.9  \scriptstyle \pm 0.0}$ & \textcolor{gray}{$100$} \tabularnewline
& Clean recall & $-$ & $\underline{81.9  \scriptstyle \pm 0.1}$ & $94.6  \scriptstyle \pm 0.4$ & $\mathbf{89.3  \scriptstyle \pm 0.1}$ & $\mathbf{89.3  \scriptstyle \pm 0.1}$ & \textcolor{gray}{$100$} \tabularnewline
& Noise recall & $-$ & $50.9  \scriptstyle \pm 0.4$ & $\underline{22.3  \scriptstyle \pm 0.4}$ & $\mathbf{80.5  \scriptstyle \pm 0.4}$ & $\mathbf{80.5  \scriptstyle \pm 0.1}$ & \textcolor{gray}{$100$} \tabularnewline
\cmidrule{3-8}
 & Accuracy & $\textcolor{gray}{56.9\scriptstyle\pm0.1}$ & $\mathbf{59.5  \scriptstyle \pm 0.1}$ & $58.9  \scriptstyle \pm 0.2$ & $58.6  \scriptstyle \pm 0.2$ & $\underline{58.0  \scriptstyle \pm 0.1}$ & \textcolor{gray}{$60.2  \scriptstyle \pm 0.2$} \tabularnewline
\midrule
\multirow{3}{*}{$0.8$} & AUROC & $-$ & $58.1  \scriptstyle \pm 0.1$ & $\underline{54.8  \scriptstyle \pm 0.2}$ & $\mathbf{63.4  \scriptstyle \pm 0.2}$ & $62.3  \scriptstyle \pm 0.1$ & \textcolor{gray}{$100$} \tabularnewline
& Clean recall & $-$ &  $\underline{86.7  \scriptstyle \pm 0.2}$ & $90.8  \scriptstyle \pm 0.1$ & $\mathbf{95.1  \scriptstyle \pm 0.4}$ & $93.4  \scriptstyle \pm 0.0$ & \textcolor{gray}{$100$} \tabularnewline
& Noise recall & $-$ & $29.5  \scriptstyle \pm 0.1$ & $\underline{18.8  \scriptstyle \pm 0.3}$ & $\mathbf{31.7  \scriptstyle \pm 0.1}$ & $31.2  \scriptstyle \pm 0.1$ & \textcolor{gray}{$100$} \tabularnewline
\cmidrule{3-8}
 & Accuracy & \textcolor{gray}{$38.5\scriptstyle\pm0.1$} & $\mathbf{45.1  \scriptstyle \pm 0.2}$ & $43.1  \scriptstyle \pm 0.2$ & $\underline{41.3  \scriptstyle \pm 0.2}$ & $41.7 \scriptstyle \pm 0.1$ & \textcolor{gray}{$46.2  \scriptstyle \pm 0.2$} \tabularnewline
\bottomrule
\end{tabular}}
\end{table*}

\begin{figure}[t]
    \centering
    \includegraphics[width=.7\linewidth]{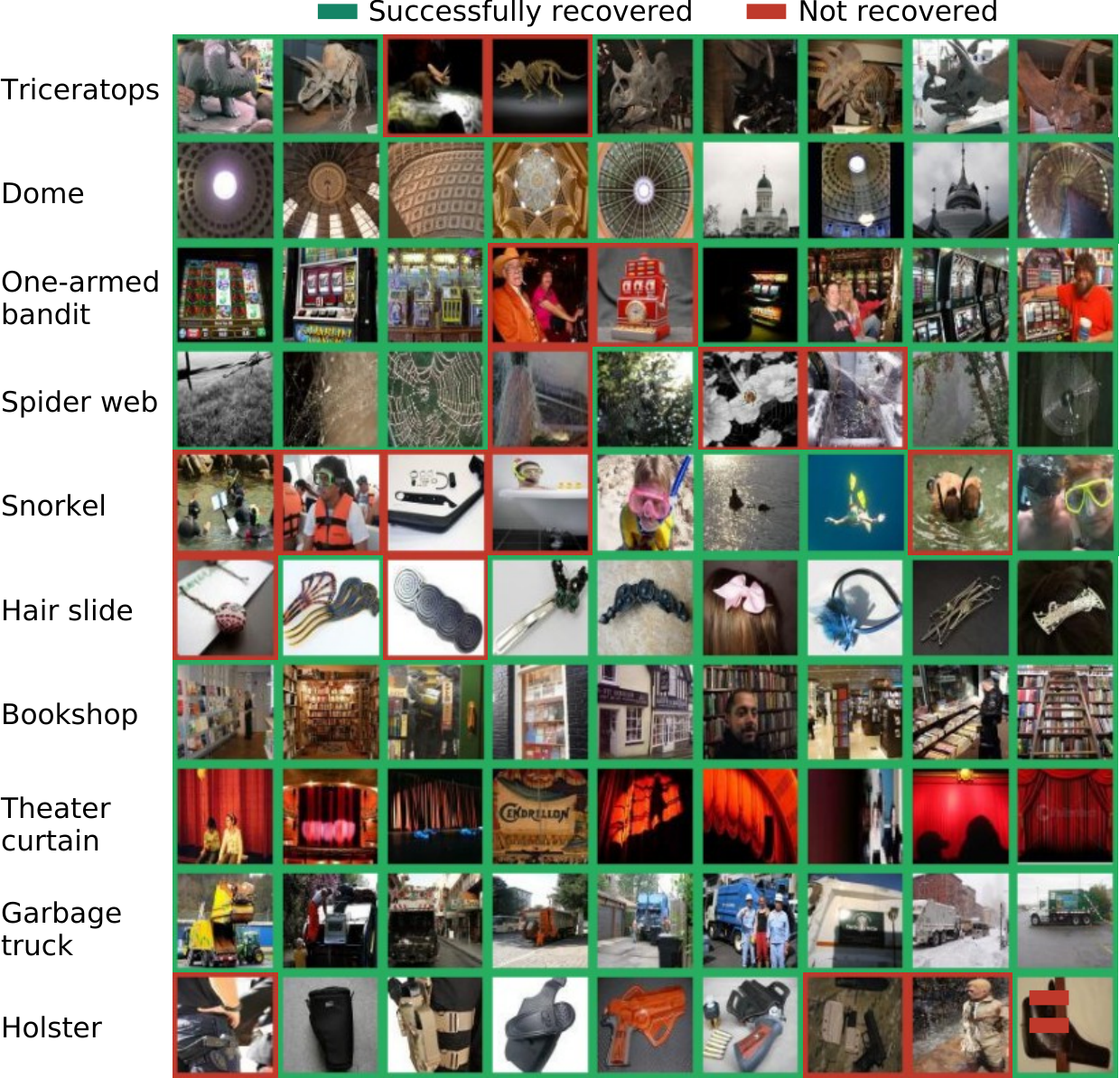}
    \caption{Examples of clean samples missed by our linear separation $W_{PLS}$ but correctly recovered (green) by a small loss approach, here PLS. $20\%$ noise CNWL. \label{fig:missedLinmain}}
    \vspace{-15pt}
\end{figure}
\begin{figure}[t]
    \centering
    \includegraphics[width=.8\linewidth]{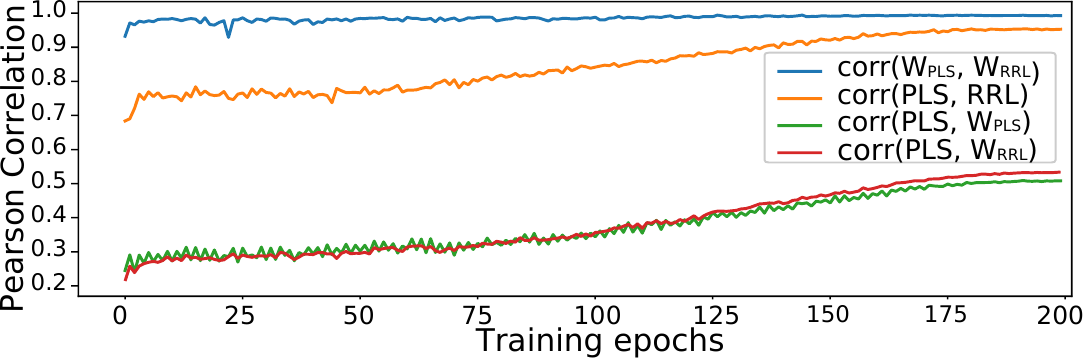}
    \caption{Low correlation of our linear separation with the PLS and RRL metrics trained on CNWL with $20\%$ web noise. $W_{PLS/RRL}$ denotes using PLS or RRL for $\hat{\mathcal{T}}$. \label{fig:decorr}}
\end{figure}

\subsection{Does better noise detection imply better classification?}
We aim to quantify the accuracy benefits of $W$ over loss-based or distance-based noise detection strategies. We select PLS~\cite{2023_WACV_PLS} for the loss-based approach (small loss strategy as in~\cite{2019_ICML_BynamicBootstrapping,2020_ICLR_DivideMix,2021_CVPR_JoSRC,2021_arXiv_propmix}) and RRL~\cite{2021_ICCV_RRL} for the distance-based approach (similar to~\cite{2021_CVPR_MOIT,2022_CVPR_NeighborConsistency,2023_PR_scanmix}). To avoid interacting with complex noise-robust mechanisms, we employ an ignore-the-noise algorithm whereby we train on the detected clean samples only using a cross-entropy loss. To obtain the RRL and PLS detection, we train the algorithms using the official code and utilize the noise detection at the end of training. We then estimate $W_{PLS}$ and $W_{RRL}$ as detailed in Section~\ref{sec:estim} using unsupervised SimCLR features. Table~\ref{tab:strongercleannoisy} reports the results on the CNWL under $20\%$ and $80\%$ web noise where we report noise detection performance by computing an AUROC curve and the clean or noisy recall as well as the classification accuracy of the ignore-the-noise algorithm.

Surprisingly, we observe that although $W_{RRL/PLS}$ improves the noise metrics in terms of AUROC and noise recall, using it to detect the noise decreases the classification accuracy of~$\Phi$. This implies that $W$ mis-identifies important samples needed to achieve high accuracy classification.

\subsection{Clean samples missed by the linear separation}
Following the observation of missing important samples in the previous section, we take a look at the clean images missed by $W$. We observe that missed clean samples predominantly represent the target object on a uniform background (typically black or white). In the context of ID and OOD separation through the alignment and uniformity of unsupervised contrastive learning presented in Section~\ref{sec:separationsncf}, this observation suggests that the unsupervised contrastive algorithm aligns uniformly colored background using this simple visual cue and independently of the ID or OOD class depicted. This problem is similar to the case where the OOD noise is structured, i.e. contains subsets of highly similar OOD images (humans holding OOD objects would be a common example). 

Although $W$ misses important examples, because these depict the target object with no distractors in the background, we suggest that they will easily be detected by the original SOTA noise detection metrics, biased toward detecting simple to fit or highly representative samples. In fact, we show in Figure~\ref{fig:missedLinmain} randomly selected clean examples missed by $W_{PLS}$ most of which are correctly retrieved by PLS. Examples for the opposite scenario can be found in the supplementary material.

\subsection{Linear Separation Alternating}
Because PLS and RRL retreive samples that $W$ misses, we aim to quantify the correlation between these noise detection metrics to justify their complementarity. We observe in Figure~\ref{fig:decorr} that while the noise detection of RRL and PLS remain correlated during training ($>0.8$ Pearson correlation) our linear separation $W$ is much more decorrelated with either RRL or PLS ($<0.5$). This low correlation further motivates the complementarity of $W$ with SOTA noise detection approaches.
We also notice that using either PLS or RRL for the trusted subset $\hat{\mathcal{T}}$ leads to very similar linear separation as $W_{PLS}$ and $W_{RRL}$ are highly correlated, explained by RRL and PLS being highly correlated to begin with.

To combine $W$ and $Z$ (PLS or RRL), we experiment with multiple combination strategies including voting or successive use (see Section~\ref{sec:fusing}). We find that alternating every epoch between $W$ and PLS or RRL to be the better strategy. One dominant advantage of the alternating strategy is that it prevents forgetting one noise-detection over the other, effectively avoiding a form of confirmation bias~\cite{2020_IJCNN_Pseudo} where mis-detections become hard to correct. We name this alternating noise detection strategy Linear Separation Alternating or LSA. Results comparing combination strategies are available in the experiments, Section~\ref{sec:fusing}.

\begin{figure}[t]
    \centering
    \includegraphics[width=\linewidth]{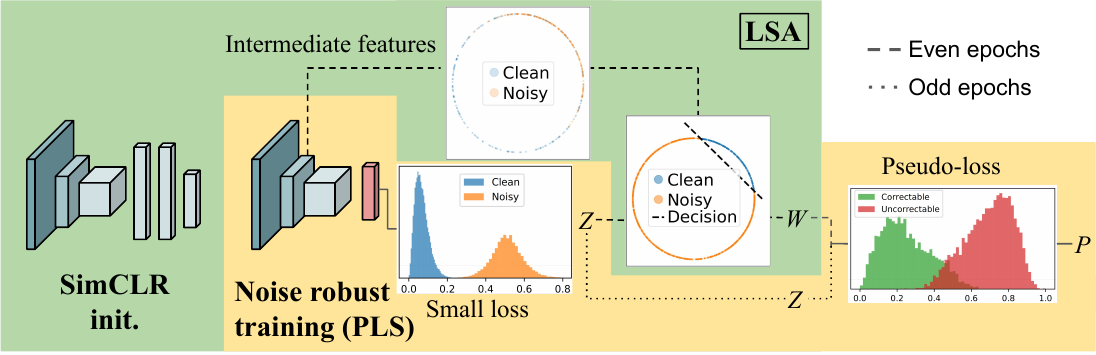}
    \caption{Illustration of the noise detection of PLS with LSA (PLS-LSA). We use $Z$ to estimate the linear separation $W$ on even epochs. \label{fig:plslsa}}
\end{figure}

\subsection{PLS-LSA}
LSA is independent from the noise-robust algorithm used whether it performs distance-based or loss-based noise detection. We choose to build on PLS~\cite{2023_WACV_PLS}, a semi-supervised strong baseline in web-noise robust algorithms. The following is a quick overview of the PLS algorithm~\cite{2023_WACV_PLS}.
In PLS, the network predicts two noisiness estimation metrics: a general noisiness $z(\bx_i, \by_i) = \bz_i$ that estimates if a sample is clean $\bz_i=1$ or noisy $\bz_i=0$ (small loss based, using a two mode gaussian mixture~\cite{2019_ICML_BynamicBootstrapping,2020_ICLR_DivideMix}) and the pseudo-loss prediction $p(\bx_i, \tilde{\by}_i, \bz_i) = \bp_i$ that estimates whether a semi-supervised imputation $\tilde{\by}_i$ is a trustworthy correction for a noisy sample ($\bp_i=1$). PLS optimizes $3$ losses : 
\begin{align}
\text{L}_{\texttt{sup}}(\bx_i, \by_i, \bz_i) = - \bz_i \times \by_i \times \log(\texttt{softmax}(\Phi(\bx_i))), \\
\text{L}_{\texttt{ssl}}(\bx_i, \tilde{\by}_i, \bp_i) = - \bp_i \times \tilde{\by}_i \times \log(\texttt{softmax}(\Phi(\bx_i)))
\end{align}
and $L_{cont}$ a supervised contrastive objective~\cite{2021_ICLR_iMix} that uses a SimCLR augmented view $\bx_i'$ and is sensitive to $\bp_i$ whose definition we refer to the original paper~\cite{2023_WACV_PLS}.
The final training loss in PLS without subscripts is 
\begin{align}
    \text{L}_{\texttt{PLS}}(\bx,\bx',\by,\bz,\bp) = \text{L}_{\texttt{sup}}(\bx, \by, \bz) + \text{L}_{\texttt{ssl}}(\bx,\tilde{\by},\bz,\bp) + \text{L}_{\texttt{cont}}(\bx, \bx', \by, \bp)   
\end{align}
We call our version of PLS using LSA, PLS-LSA where we pretrain $\Phi$ using SimCLR and replace $Z$ with $W$ on even epochs, i.e. on even epochs we compute $\bp_i = p(\bx_i, \tilde{\by}_i, w_i)$ and $L_{sup}(\bx_i, \by_i, w_i)$. For $W$ we use features extracted in the second ResNet block, an ablation can be found in the supplementary material.

\subsection{Semi-supervised imputation and Co-training~\label{sec:imput}}
Because PLS-LSA lacks some common additions to the recent noise-robust literature, we propose to use a stronger data augmentation for semi-supervised imputation and introduce an optional voting co-training strategy. 
We modify the PLS label imputation strategy as follows: given $\bx_i''$ augmented using RandAugment~\cite{2020_CVPRW_randaugment}, we modify the semi-supervised loss of PLS to
\begin{align}
\text{L}_{\texttt{ssl}}(\bx_i, \bx_i'', \tilde{\by}_i, \bp_i) = - \bp_i \times \texttt{sg}(\texttt{softmax}(\Phi(\bx_i))) \times \log(\texttt{softmax}(\Phi(\bx_i'')))
\end{align} where $\texttt{sg}(.)$ is the stop gradient operation.
This imputation strategy is in line with recent semi-supervised classification research~\cite{2020_arXiv_FixMatch,2021_NeurIPS_flexmatch}.

PLS-LSA+ is a co-training strategy for PLS-LSA that uses two co-trained networks. We use a voting approach where the two networks vote for noisy samples detection~$\bz_i$, $w_i$ and $\bp_i$ as well as for classification at test time. Our voting noise detection is different from previous approaches~\cite{2020_ICLR_DivideMix,2023_PR_scanmix,2018_NeurIPS_CoTeaching} where networks predict noisiness for each other.
Additionally, before voting on~$\bp_i$ we introduce a co-guessing strategy where a semi-supervised prediction $\tilde{\by}_i$ of network 1 is evaluated as correct by network 2. The naive strategy would be each network evaluating if their own guess is correct which introduces more confirmation bias.

%% file: sec/4_exp.tex
\section{Experiments\label{sec:experiments}}

\subsection{Structure of the experiments section}
We structure the experiment section as follows.
%First, we study how to accurately estimate the linear separation between ID and OOD given a inital detection $Z$ starting from the noise detection metric of PLS or RRL up to using a trusted ID/OOD subset labeled manually.
%Then we show how, despite being more specific and sensitive than the metrics of PLS and RRL, our linear separation noise detection does not produce a more accurate classification algorithm. We observe that this low performance is due to few samples that are miss-identified by our linear separation yet correctly retreived by the metrics of PLS and RRL. This motivates a complimentary use of the noise detection metrics. 
First we study different combination strategies for PLS and $W_{PLS}$. We then conduct an ablation study of PLS-LSA to highlight the importance of each of our proposed addition over PLS.
We finally compare PLS-LSA with SOTA algorithms on the Controlled Noisy Web Labels (CNWL) dataset~\cite{2018_ICML_MentorNet} and real world datasets mini-Webvision~\cite{2017_arXiv_WebVision} and Webly-fg~\cite{2021_ICCV_weblyfinegrained}.
The CNWL dataset corrupts miniImageNet~\cite{2016_NIPS_MiniImageNet} with human curated web-noisy examples. The dataset proposes noise ratios ranging from $20$ to $80\%$. Following previous research, we train on the CNWL at a resolution of $32^2$ using a PreActivation ResNet18~\cite{2016_ECCV_preactresnet}. mini-WebVision is a subset of the first $50$ classes of Webvision~\cite{2017_arXiv_WebVision} which mimic ImageNet~\cite{2012_NeurIPS_ImageNet} classes. We train on mini-WebVision at a resolution of $224^2$ using an InceptionResNetV2~\cite{2016_AAAI_Inception}. The Webly-fg datasets are noisy datasets that target fined-grained classification of aircrafts, birds or cars. We train at a resolution of $448^2$ using a ResNet50 initialized either on ImageNet as done in previous research or using SimCLR. All datasets contain unidentified web-noisy samples, which are either OOD or ID noisy (mislabeled). We compare the performance of noise-robust algorithms trained on web-noisy datasets by their ability to accurately classify a clean validation set. More detailed experimental settings are available in the supplementary material.

Our experimental settings are the same as used in PLS~\cite{2023_WACV_PLS} and comparable to evaluation settings used in the algorithms we compare with. Unless otherwise specified, we initialize our networks using SimCLR~\cite{2020_ICML_SimCLR} and solo-learn~\cite{2022_JMLR_sololearn}.

\begin{table}[t]
\centering{}\caption{Best strategy to combine PLS and $W_{PLS}$ on the CNWL.\label{tab:alternate}}
\global\long\def\arraystretch{0.9}%
\resizebox{.6\linewidth}{!}{{{}}
\begin{tabular}{c>{\centering}c>{\centering}c>{\centering}c>{\centering}c>{\centering}c>{\centering}c>{\centering}c>{\centering}c>{\centering}c>{\centering}c}
\toprule
Noise ratio & 0.2 & 0.4 & 0.6 & 0.8 \tabularnewline
\midrule
PLS & $62.30  \scriptstyle \pm 0.24$ & $59.11  \scriptstyle \pm 0.28$ & $54.26  \scriptstyle \pm 0.20$ & $48.71  \scriptstyle \pm 0.36$ \tabularnewline
$W_{PLS}$ & $62.97  \scriptstyle \pm 0.23$ & $60.41  \scriptstyle \pm 0.05$ & $52.18  \scriptstyle \pm 0.12$ & $47.11  \scriptstyle \pm 0.08$ \tabularnewline 
AND & $62.66  \scriptstyle \pm 0.34$ & $58.80  \scriptstyle \pm 0.41$ & $54.82  \scriptstyle \pm 0.15$ & $48.19  \scriptstyle \pm 0.11$ \tabularnewline 
OR & $58.79  \scriptstyle \pm 0.40$ & $58.27  \scriptstyle \pm 0.34$ & $50.57  \scriptstyle \pm 1.11$ & $45.93  \scriptstyle \pm 0.69$ \tabularnewline 
$W_{PLS}\xrightarrow{}$PLS & $62.88  \scriptstyle \pm 0.46$ & $59.89  \scriptstyle \pm 0.14$ & $52.34  \scriptstyle \pm 0.12$ & $48.88  \scriptstyle \pm 0.23$ \tabularnewline 
PLS$\xrightarrow{}W_{PLS}$ & $63.49  \scriptstyle \pm 0.12$ & $60.21  \scriptstyle \pm 0.12$ & $55.36  \scriptstyle \pm 0.84$ & $49.23  \scriptstyle \pm 0.21$ \tabularnewline 
LSA & $\mathbf{64.20}  \scriptstyle \pm 0.16$ & $\mathbf{60.98}  \scriptstyle \pm 0.24$ & $\mathbf{55.64}  \scriptstyle \pm 0.30$ & $\mathbf{49.73}  \scriptstyle \pm 0.13$ \tabularnewline 
\midrule
\textcolor{gray}{Oracle} & \textcolor{gray}{$64.10  \scriptstyle \pm 0.10$} & \textcolor{gray}{$61.45  \scriptstyle \pm 0.22$} & \textcolor{gray}{$56.04  \scriptstyle \pm 0.39$} & \textcolor{gray}{$50.19  \scriptstyle \pm 0.48$} \tabularnewline 
\bottomrule
\end{tabular}}
\end{table}

\subsection{Combining PLS and $W_{PLS}$\label{sec:fusing}}
We investigate here mulitple strategies for PLS-LSA combining the decorrelated $W$ and PLS so that we can maximize classification accuracy on the held out validation set. We propose to use AND or OR logic operators (clean is false and noisy true), sucessive noise detection where we train using either metric for the first half of training and then switch to the other for the remainder ($W_{PLS}\xrightarrow{}$PLS and PLS$\xrightarrow{}W_{PLS}$) or our alternating approach (LSA) where either metric is alternatively used every epoch. Table~\ref{tab:alternate} displays our results. 

We find that two strategies are superior to the PLS baseline: the second best strategy is PLS$\xrightarrow{}W_{PLS}$, explained because the simple samples $W_{PLS}$ misses are less important to get right in later training steps when the network has already learned strong base features for each class. The LSA strategy is the best approach overall, we believe this is because training the algorithm on both detection regularly allows to learn from the clean training examples provided by both metrics while avoiding over-fitting either metric's defects. These results solidify LSA as the better alternative for combining $W$ and PLS.

\begin{table}[t]
\parbox{.55\linewidth}{
\centering{}\caption{Ablation study CNWL~\label{tab:abla}}
\global\long\def\arraystretch{0.9}%
\resizebox{\linewidth}{!}{{{}}%
\begin{tabular}{l>{\centering}c>{\centering}c>{\centering}c>{\centering}c>{\centering}c}
\toprule
 Dataset & mini $20\%$ & mini $80\%$ \tabularnewline
 \midrule
 \multicolumn{4}{c}{Baselines} \tabularnewline
 \midrule
 mixup & $57.27  \scriptstyle \pm 0.39$ & $38.48  \scriptstyle \pm 0.24$ \tabularnewline
 mixup + SimCLR & $57.03  \scriptstyle \pm 0.10$ & $39.62  \scriptstyle \pm 0.28$ \tabularnewline
 PLS & $62.83  \scriptstyle \pm 0.39$ & $45.80  \scriptstyle \pm 0.72$ \tabularnewline 
 PLS + SimCLR & $62.39  \scriptstyle \pm 0.14$ & $47.21  \scriptstyle \pm 0.63$ \tabularnewline 
 PLS ours & $63.25  \scriptstyle \pm 0.24$ &  $48.03  \scriptstyle \pm 0.38$ \tabularnewline
 \midrule
 \multicolumn{4}{c}{PLS-LSA ablation} \tabularnewline
 \midrule
 PLS-LSA &  $64.61  \scriptstyle \pm 0.51$ & $48.20  \scriptstyle \pm 0.16$ \tabularnewline 
 PLS-LSA no SimCLR &  $64.04  \scriptstyle \pm 0.27$ & $47.29  \scriptstyle \pm 0.27$ & \tabularnewline  
 PLS-LSA no DA & $63.25  \scriptstyle \pm 0.21$ & $43.55  \scriptstyle \pm 0.45$ \tabularnewline 
 PLS-LSA no SimCLR DA & $61.83  \scriptstyle \pm 0.55$ & $43.75  \scriptstyle \pm 0.35$ & \tabularnewline 
 \midrule
 \multicolumn{4}{c}{PLS-LSA+ ablation} \tabularnewline
 \midrule
 PLS-LSA+ &  $66.52  \scriptstyle \pm 0.10$ & $52.03  \scriptstyle \pm 0.32$ \tabularnewline 
 %PLS-LSA+ no DA &  &  \tabularnewline 
 PLS-LSA+ no SimCLR DA & $66.34  \scriptstyle \pm 0.27$  & $47.68  \scriptstyle \pm 0.59$ \tabularnewline 
 \bottomrule
\end{tabular}}}
\hfill
\parbox{.35\linewidth}{
\centering{}\caption{Ablation study Webvision~\label{tab:ablawebvis}}
\global\long\def\arraystretch{0.9}%
\resizebox{\linewidth}{!}{{{}}%
\begin{tabular}{l>{\centering}c>{\centering}c>{\centering}c>{\centering}c>{\centering}c}
\toprule
 Algorithm & Webvision \tabularnewline
 \midrule
 mixup & $77.99$ \tabularnewline
 mixup + SimCLR & $78.88$ \tabularnewline
 PLS & $79.01$ \tabularnewline 
 PLS-LSA & $81.36$ \tabularnewline 
 PLS-LSA no SimCLR  & $78.68$ \tabularnewline  
 PLS-LSA no DA  & $79.00$ \tabularnewline 
 PLS-LSA no SimCLR DA & $76.80$ \tabularnewline

 \bottomrule
\end{tabular}}}
\vspace{-10pt}
\end{table}

\subsection{Ablation study\label{sec:abla}}

We conduct an ablation study to evaluate the importance of each of our design choices in Table~\ref{tab:abla}. We first ablate on the CNWL dataset under $20\%$ and $80\%$ noise. We evaluate the improvements of SimCLR when added to a simple noise robust training using Mixup or the original PLS algorithm. Interestingly we observe that unsupervised initialization has little effect on validation accuracy for lower noise ratios. We also report PLS (ours) which denotes our improved version of PLS (PLS-LSA without LSA) which uses SimCLR initialization and improved data augmentations. Our version performs slightly better when compared to the original PLS.
The second part of the table ablates elements from PLS-LSA: SimCLR initialization, stronger data augmentation (DA) or both (nothing). Strong data augmentations appears to be an important element of PLS-LSA. This is explained by our semi-supervised imputation strategy being largely dependent on stronger data augmentations whereas another SSL imputation strategie (i.e. MixMatch~\cite{2019_NIPS_MixMatch} used in PLS) would be better suited when not having access to stronger DA.
Interestingly, PLS-LSA does not catastrophically fail when we remove the SimCLR initialization. This hints towards observing the linear separation without self-supervised pre-training and shows that the alternating strategy provides stability though to the original PLS detection.
We finally observe that PLS-LSA+ nothing manages to use co-training to maintain a high accuracy in the lower noise scenario even if we remove SimCLR initialization and strong DA.

We additionally run ablations experiments on mini-Webvision to measure impacts in the real world. Results are available in Table~\ref{tab:ablawebvis}. In this context, SimCLR initialization appears to play a more important role than on the CNWL and is important to maintain a good classification accuracy with PLS-LSA.

\begin{table*}[t]
    \caption{CNWL~\cite{2020_ICML_MentorMix} ($32\times 32$). We run PLS and PLS-LSA; other results are from~\cite{2023_WACV_PLS}. We report top-1 best accuracy and bold the best results with and without co-training. Accuracy results averaged over 3 random seeds $\pm$ one std.
    \label{tab:sotamini32}}
    \global\long\def\arraystretch{0.9}%
    \centering
    \resizebox{\textwidth}{!}{%
    \centering
    \begin{tabular}{c>{\centering}c>{\centering}c>{\centering}c>{\centering}c>{\centering}c>{\centering}c>{\centering}c>{\centering}c>{\centering}c>{\centering}c>{\centering}c>{\centering}c}
    \toprule
    & \multicolumn{6}{c}{No co-training} & \multicolumn{6}{c}{Co-training} \tabularnewline
    \cmidrule(lr){2-7} \cmidrule(lr){8-13}
    Noise level & M & MM & FaMUS & SNCF & PLS & PLS-LSA & DM & SM & PM  & LRM & MDM  & PLS-LSA+ \tabularnewline
    \midrule  
    20 &  $49.10$ & $51.02$ & $51.42$ & $61.56$ & $63.25  \scriptstyle \pm 0.24$ & $\mathbf{64.43}  \scriptstyle \pm 0.21$ & $50.96$ & $59.06$ & $61.24$ & $56.03  \scriptstyle \pm 0.5$ & $64.40$ &  $\mathbf{66.52}  \scriptstyle \pm 0.10$ \tabularnewline
    40 & $46.40$ & $47.14$ & $48.03$ & $59.94$ & $60.42  \scriptstyle \pm 0.23$ & $\mathbf{61.14}  \scriptstyle \pm 0.35$ & $46.72$  & $54.54$ & $56.22$ & $50.69  \scriptstyle \pm 0.3$ & $61.40$ &   $\mathbf{63.42}  \scriptstyle \pm 0.42$ \tabularnewline
    60 & $40.58$ & $43.80$ & $45.10$ & $54.92$ & $55.34  \scriptstyle \pm 0.38$ & $\mathbf{57.18}  \scriptstyle \pm 0.30$ & $43.14$ & $52.36$ &  $52.84$ & $46.81  \scriptstyle \pm 0.3$ & $56.20$  & $\mathbf{59.41}  \scriptstyle \pm 0.30$ \tabularnewline
    80 & $33.58$ & $33.46$ & $35.50$ & $45.62$ & $48.03  \scriptstyle \pm 0.20$ & $\mathbf{49.53}  \scriptstyle \pm 0.46$ & $34.50$ & $40.00$ & $43.42$ & $38.24  \scriptstyle \pm 0.2$ & $47.80$ & $\mathbf{52.03}  \scriptstyle \pm 0.32$ \tabularnewline
       \bottomrule
    \end{tabular}}
\end{table*}

\begin{table*}[t]
    \caption{Classification accuracy for training on mini-Webvision using InceptionResNetV2. We denote with $\dagger$ algorithms using unsupervised initialization. We test on the mini-Webvision valset and ImageNet 1k test set (ILSVRC12).  We run PLS and PLS-LSA, other results are from SNCF~\cite{2022_ECCV_SNCF}. We bold the best results. Accuracy results averaged over 3 random seeds $\pm$ one std.\label{tab:webvis}}
    \centering
    \global\long\def\arraystretch{0.9}%
    \resizebox{\textwidth}{!}{{{}}%
    \begin{tabular}{l>{\centering}c>{\centering}c>{\centering}c>{\centering}c>{\centering}c>{\centering}c>{\centering}c>{\centering}c>{\centering}c>{\centering}c>{\centering}c>{\centering}c>{\centering}c>{\centering}c>{\centering}c}
    \toprule
     & & \multicolumn{6}{c}{No co-training} & \multicolumn{8}{c}{Co-training} \tabularnewline
     \cmidrule(lr){3-9} \cmidrule(lr){10-16}
     
    Testset & & M & MM & RRL & FaMUS & PLS & FLY & $\dagger$PLS-LSA & DM & ELR+ & DSOS &  $\dagger$SM  & RM & SNCF+ & $\dagger$PLS-LSA+ \tabularnewline
    \midrule
    \multirow{2}{*}{mini-WebVision} & top-1 & $75.44$ & $76.0$ & $77.80$ & $79.40$ & $79.01  \scriptstyle \pm 0.33$ & $80.96$ & $\mathbf{81.28}  \scriptstyle \pm 0.11$ & $77.32$ & $77.78$  & $78.76$ & $80.04$ & $79.91$ & $80.24$ & $\mathbf{82.08}  \scriptstyle \pm 0.28$ \tabularnewline
    & top-5 & $90.12$ & $90.2$ & $91.30$ & $92.80$ & $92.05  \scriptstyle \pm 0.46$ & $93.56$ &$\mathbf{94.12}  \scriptstyle \pm 0.09$ & $91.64$ & $91.68$ & $92.32$ & $93.04$ & $93.61$ & $93.44$ & $\mathbf{94.16}  \scriptstyle \pm 0.10$\tabularnewline
    \multirow{2}{*}{ILSVRC12} & top-1 & $71.44$ & $72.9$ & $74.40$ & $77.00$ & $76.15  \scriptstyle \pm 0.35$ & $--$ & $\mathbf{78.32}  \scriptstyle \pm 0.74$ & $75.20$ & $70.29$ & $75.88$ & $75.76$ & $77.39$ & $77.12$ &  $\mathbf{79.19}  \scriptstyle \pm 0.52$ \tabularnewline
    & top-5 & $89.40$ & $91.10$ & $90.90$ & $92.76$ & $92.53  \scriptstyle \pm 0.23$ & $--$ & $\mathbf{94.64}  \scriptstyle \pm 0.20$ & $90.84$ & $89.76$ & $92.36$&  $92.60$ & $94.26$ & $94.32$ & $\mathbf{94.84}  \scriptstyle \pm 0.25$\tabularnewline
       \bottomrule
    \end{tabular}}
     
\end{table*}

\subsection{SOTA comparison on the CNWL dataset}
We compare with related SOTA on the CNWL dataset corrupted with $20, 40, 60, 80$\% web noise in Table~\ref{tab:sotamini32}. We report the accuracy results of both PLS-LSA and PLS-LSA+. The noise-robust algorithms we compare with are Mixup~\cite{2018_ICLR_Mixup} a noise robust regularization, FaMUS~\cite{2021_CVPR_FaMUS} a meta learning approach and sample correction algorithms: DivideMix (DM)~\cite{2020_ICLR_DivideMix}, MentorMix (MM)~\cite{2020_ICML_MentorMix}, ScanMix~\cite{2023_PR_scanmix}, PropMix (PM)~\cite{2021_arXiv_propmix}, SNCF~\cite{2022_ECCV_SNCF}, LongReMix (LRM)~\cite{2023_PR_longremix}, Manifold DivideMix (MDM)~\cite{2023_arXiv_MDM} and PLS~\cite{2023_WACV_PLS}. We find that PLS-LSA improves over existing approaches even when these use a co-training strategy (PLS-LSA only uses one network). PLS-LSA+ further improves the classification accuracy by $2$ to $5$ absolute points across noise levels.

\subsection{Real world datasets\label{sec:realworldexp}}
We now evaluate PLS-LSA on real world datasets. For mini-Webvision we add to the comparison a robust loss algorithm Early Learning Regularization (ELR)~\cite{2020_NeurIPS_EarlyReg}, as well as additional sample correction algorithms: Robust Representation Learning (RRL)~\cite{2021_ICCV_RRL}, DSOS~\cite{2022_WACV_DSOS}, RankMatch (RM)~\cite{2023_ICCV_rankmatch} and LNL-Flywheel (FLY)~\cite{2024_arxiv_Fly}. We also report our results on the webly fine-grained datasets as well as for Co-teaching~\cite{2018_NeurIPS_CoTeaching}, PENCIL~\cite{2019_CVPR_PENCIL}, SELFIE~\cite{2019_ICML_SELFIE}, Peer-learning~\cite{2021_ICCV_weblyfinegrained} and Progressive Label Correction (PLC)~\cite{2021_ICLR_featurenoise} which are all sample correction algorithms.
\vspace{-10pt}
\subsubsection{mini-Webvision}
We train PLS-LSA on mini-Webvision and report test results on the validation set of mini-Webvision and well as on the validation set of ImageNet2012~\cite{2012_NeurIPS_ImageNet} in Table~\ref{tab:webvis}. We outperform co-training methods with PLS-LSA using only one network and PLS-LSA+ sets a new state-of-the-art by improving over PLS-LSA in terms of top-1 accuracy but we notice no significant improvements for top-5 accuracy. We report additional results when training a ResNet50 on mini-Webvision in the supplementary material where we observe similar improvements of PLS-LSA and PLS-LSA+ when compared to related works.
\vspace{-10pt}
\subsubsection{Webly-fg datasets}
We train PLS-LSA on the Webly-fg datasets~\cite{2021_ICCV_weblyfinegrained} that present the added challenge of fine-grained classification over mini-Webvision. We report results on the bird, car and aircraft subsets in Table~\ref{tab:sotawebfg}. Because other methods use ImageNet weights for pre-training, we report results using either ImageNet or SimCLR pretraining to exhibit the linear separation between ID and OOD noise. We find that PLS-LSA only marginally improves over PLS even in the case where we use self-superivsed features. We found that learning strong SimCLR features for Webly-fg datasets is challenging due to the fine grained nature of the dataset. It could be the case that using a different set of data augmentations or a different self-supervised algorithm would help improve our performance further. PLS-LSA+ improves $0.7$ to $0.8$ points over PLS-LSA.

\begin{table}[t]
    \centering
    \caption{Comparison against state-of-the-art algorithms on the fine grained web datasets,  we run PLS-LSA and bold the best results. Results for other algorithms from~\cite{2023_WACV_PLS}. Top-1 best accuracy. \label{tab:sotawebfg}}
    \global\long\def\arraystretch{0.8}%
    \resizebox{.65\linewidth}{!}{{{}}%
    \begin{tabular}{l>{\centering}c>{\centering}c>{\centering}c>{\centering}c>{\centering}c}
    \toprule
      Initialization & Algorithm & Web-Aircraft & Web-bird & Web-car \tabularnewline
       \midrule
    \multirow{10}{*}{ImageNet} & CE & 60.80 & 64.40 & 60.60 \tabularnewline
    & Co-teaching & 79.54 & 76.68 & 84.95 \tabularnewline
    & PENCIL & 78.82 & 75.09 & 81.68 \tabularnewline
    & SELFIE & 79.27 & 77.20 & 82.90 \tabularnewline 
    & DivideMix & 82.48 & 74.40 & 84.27 \tabularnewline
    & Peer-learning & 78.64 & 75.37 & 82.48 \tabularnewline
    & PLC & 79.24 & 76.22 & 81.87 \tabularnewline
       %Jo-SRC & 82.73 & 81.22 & 88.13 \tabularnewline
       %$\star$Co-LDL & 83.83 & 81.02 & 89.17 \tabularnewline
       %$\star$PNP & 85.54 & 81.93 & 90.11 \tabularnewline
    & PLS & 87.58 & 79.00 & 86.27 \tabularnewline
    %\cmidrule{2-5}
    & PLS-LSA & 87.70 & 79.20 & 86.58\tabularnewline
    & PLS-LSA+ & 88.42 & 79.77 & 87.24 \tabularnewline
    \midrule
    \multirow{2}{*}{SimCLR} & PLS-LSA & 87.82 & 79.47 & 86.76\tabularnewline
    & PLS-LSA+ & \textbf{88.51} & \textbf{80.03} & \textbf{87.50}  \tabularnewline
    \bottomrule
    \end{tabular}}
    \vspace{-10pt}
\end{table}

%% file: sec/5_conclu.tex
\section{Conclusion}
This paper builds on the previously observed linear separation of ID and OOD images in unsupervised contrastive feature spaces in the context of label noise datasets. We observe that the linear separation of ID and OOD features is not as evident as previously observed when moving to real-world data yet becomes apparent again when looking at lower level features. Instead of relying on clustering as done in previous research, we propose to compute the linear separation using an approximate ID/OOD detection using state-of-the-art noise-robust metrics. Although we find our noise detector to be highly accurate, we do not observe classification accuracy gains when compared to less accurate SOTA noise detectors. We evidence that the few samples we mis-identify are crucial to train a strong classifier. We combine our detection together with PLS by alternating the noise detection approach every epoch to create PLS-LSA. We further develop a co-train schedule using two networks to produce PLS-LSA+. Our results improve the SOTA classification accuracy on real-world web noise datasets.
Because we only empirically observe the linear separation in earlier layers, we stress the need for further theoretical analysis of the phenomenon and encourage further research in this direction. Other future work we recommend is to study if intelligent alternating strategies could be developed to combine both detection approaches based on the current noise detection bias in the network. We also suggest that further attention be given to whether the linear separation can be enforced from a random initialization and as training progresses to remove the need for pretraining.
\vspace{-10pt}
\subsubsection*{Acknowledgments} This publication has emanated from research conducted with the joint financial support of the Center for Augmented Reasoning (CAR) and Science Foundation Ireland (SFI) under grant number SFI/12/RC/2289\_P2. The authors additionally acknowledge the Irish Centre for High-End Computing (ICHEC) for the provision of computational facilities and support.
The authors would like to issue special remembrance to our dearly missed friend and colleague Kevin McGuinness for his invaluable contributions to our research.

%% file: sec/X_suppl.tex
\clearpage
\setcounter{page}{1}
\title{Supplementary material PLS-LSA}
\maketitle

\textbf{Supplementary material overview.} Section~\ref{sec:traindetails} details the training hyper-parameters for experiments on the CNWL, mini-Webvision and Webly-fg datasets. 
Section~\ref{sec:webvisres50} reports results when training PLS-LSA and PLS-LSA+ on mini-Webivision using a ResNet50 as well as results for related state-of-the-art algorithms. 
Section~\ref{sec:cotraining} studies if PLS-LSA+ and other SOTA co-training alternatives produce individual neural networks that are significantly more accurate than non co-trained strategies.
Section~\ref{sec:stratslsa} studies different strategies for LSA including using a trusted ID/OOD subset, different alternating strategies and computing the linear separation using features at different depth in the network. 
Section~\ref{sec:missedimp} shows the complementarity of the noise retrieval metrics used in PLS-LSA and examples of missed clean examples by one metric but retreived by the other. 
Section~\ref{sec:topnwebfg} reports top-2 and top-5 accuracy results for the Webly-fg datasets.
Finally, Section~\ref{sec:noisyimages} displays noisy images detected using PLS-LSA for the mini-Webivision or the Webly-fg datasets.

%then show examples of clean training samples missed by either our linear separation or the original PLS metric and how clean samples missed by one metric is partially recovered by the other. Section~\ref{sec:complmissed} provides further observations on the complementarity of the noise detection of PLS-LSA where we observe that both metrics are weakly correlated throughout training and show how much of the missed samples are alternatively retrieved. Section~\ref{sec:improving} show how training PLS-LSA improves the small loss noise detection of PLS. Section~\ref{sec:trusted} trains the PLS-LSA algorithm on the CNWL using trusted subsets of $100$, $1.000$ and $10.000$ ID/OOD samples to estimate $W$ where we observe small improvements over the unsupervised PLS-LSA especially for higher noise ratios. Sections~\ref{sec:alternstrat}, \ref{sec:linsepdepth} and \ref{sec:pen} respectively study different alternating strategies for PLS-LSA, training PLS-LSA computing the linear separation using features at different depth in the network and comparing our class imbalance penalty against the one used in PLS. We finally show examples of detected noisy sample for the mini-Webvision dataset in Section~\ref{sec:noisywebvis}.

\section{Training details\label{sec:traindetails}}
\subsection{CNWL} 
The Controlled Web Noisy Label (CNWL)~\cite{2020_ICML_MentorMix} proposes a controlled web-noise corruption of MiniImageNet~\cite{2016_NIPS_MiniImageNet} where some images of the original dataset are replaced with human curated incorrect samples obtained from web-queries on the corresponding class. We train on the CNWL at a resolution of $32\times 32$ using a pre-activation ResNet18~\cite{2016_ECCV_preactresnet}. We train for $200$ epochs using a cosine decay scheduling from a learning rate of $0.1$. We optimize the network using stochastic gradient descent (SGD) with a weight decay of $0.0005$. For training augmentations, we use random cropping and horizontal flipping, for the strong augmentations, we use a random resize copping strategy followed by RandAugment~\cite{2020_CVPRW_randaugment} with parameters 1 and 6. For unsupervised pretraining, we train SimCLR for $1.000$ epochs using the solo-learn~\cite{2022_JMLR_sololearn} library.

\subsection{mini-Webvision}
Webvision~\cite{2017_arXiv_WebVision} is a real world classification web-dataset over the classes of ImageNet~\cite{2012_NeurIPS_ImageNet} the original paper estimates the noise level in Webvision to be between $20\%$ to $34\%$. As in previous research, we train on the first $50$ classes (mini-Webvision) which yields $65,944$ training images and using an InceptionResNetV2~\cite{2016_AAAI_Inception} or a ResNet50~\cite{2016_CVPR_ResNet} architecture. We train at a resolution of $224 \times 224$ for $130$ epochs and otherwise the same optimization regime as for the CNWL dataset (cosine lr decay, SGD, weight decay $0.0005$) but with a batch size of $64$ and from an initial learning rate of $0.02$ ($0.01$ for ResNet50). The training augmentations are resizing to $256 \times 256$ before random cropping to $224 \times 224$ and random horizontal flipping. The strong augmentations are first resizing to $256 \times 256$ then random resize cropping to $224 \times 224$ and applying RandAugment with parameters 1 and 4. For unsupervised pretraining, we train SimCLR for $400$ epochs using the solo-learn library.

\subsection{Webly-fg}

We also evaluate PLS-LSA on the Webly-fine-grained (Webly-fg) datasets~\cite{2021_ICCV_weblyfinegrained} which are real world fine-grained classification datasets build from web queries. We train specifically on the web-bird, web-car and web-aircraft subsets that respectively contain $200$, $196$ and $100$ classes. Each dataset contains $18.388$, $21.448$, $13.503$ training, and $5.794$, $8.041$, $3.333$ test images.
We train a ResNet50 network with a batch size of $32$, at a resolution of $448 \times 448$ for $110$ epochs. Our intial learning rate is $0.006$ and we train using cosine decay, SGD and a weight decay of $0.001$). The training augmentations are resizing to $512 \times 512$ then cropping to $448 \times 448$ and random horizontal flipping. For the strong augmentations, we resize to $512 \times 512$ then random resize crop to $448 \times 448$ and apply RandAugment with parameters 1 and 4. Unsupervised pretraining is the same as Webvision but at a resolution of $448 \times 448$.

\iffalse
\begin{table}[t]
\centering{}\caption{Ablation study~\label{tab:ablaold}}

\global\long\def\arraystretch{0.9}%
\resizebox{\linewidth}{!}{{{}}%
\begin{tabular}{l>{\centering}c>{\centering}c>{\centering}c>{\centering}c>{\centering}c}
\toprule
  Dataset & mini $20\%$ & mini $80\%$ & Webvision \tabularnewline
 \midrule
 \multicolumn{4}{c}{No unsup init.} \tabularnewline
 \midrule
 mixup & $58.99 \pm 0.29$ & $33.69 \pm 3.78$ & $77.99 \pm 0.27$ 
 \tabularnewline
 PLS & $63.10 \pm 0.14$ & $46.51 \pm 0.20$ & $79.01 \pm 0.33$ \tabularnewline   
 PLS-LSA &  $64.73 \pm 0.29$ & $49.53 \pm 0.46$ & $80.61 \pm 0.11$ \tabularnewline   
 Co-train PLS + DA & $66.64 \pm 0.11$ & 2,3 & & \tabularnewline   
 Co-train PLS-LSA & $66.39 \pm 0.26$ & $52.03 \pm 0.32$ & $81.61 \pm 0.28$\tabularnewline
 \midrule
 \multicolumn{4}{c}{No unsup. init.} \tabularnewline
 \midrule
 PLS-LSA & $64.59 \pm 0.16$ & $47.60 \pm 0.72$ & $78.60 \pm 0.26$ \tabularnewline
 Co-train PLS-LSA & $66.70 \pm 0.42$ & $47.60 \pm 0.72$ & \tabularnewline
 \midrule
 \multicolumn{4}{c}{No unsup. init. and weaker DA} \tabularnewline
 \midrule
 PLS & $63.10 \pm 0.14$ & $46.51 \pm 0.20$ & $78.04 \pm 0.25$ \tabularnewline
 PLS-LSA &  $63.88 \pm 0.31$ & $45.59 \pm 0.16$ & $78.39 \pm 0.45$ \tabularnewline
 Co-train PLS-LSA & $66.36 \pm 0.14$ & 3 & \tabularnewline
 \bottomrule
\end{tabular}}
\end{table}
\fi

\begin{table*}[t]
    \caption{Classification accuracy training on mini-Webvision using ResNet50. We denote with $\dagger$ algorithms using unsupervised initialization. We test on the mini-Webvision valset and ImageNet 1k test set (ILSVRC12).  We run PLS and PLS-LSA, other results are from the respective papers. $--$ denotes that the papers did not report any results. We bold the best results. Accuracy results averaged over 3 random seeds $\pm$ one std.\label{tab:webvisres50}}
    \centering
    \global\long\def\arraystretch{0.9}%
    \resizebox{\textwidth}{!}{{{}}%
    \begin{tabular}{l>{\centering}c>{\centering}c>{\centering}c>{\centering}c>{\centering}c>{\centering}c>{\centering}c>{\centering}c>{\centering}c>{\centering}c>{\centering}c>{\centering}c>{\centering}c>{\centering}c>{\centering}c}
    \toprule       
    Valset & & M & DM & $\dagger$C2D & TCL & LCI & NCR & $\dagger$PLS-LSA & $\dagger$PLS-LSA+ \tabularnewline
    \midrule
    \multirow{2}{*}{mini-WebVision} & top-1 & $76.0\scriptstyle\pm0.2$ & $76.3\scriptstyle\pm0.36$ & $79.4\scriptstyle\pm0.3$ & $79.1$ & $80.0$ & $80.5$ & $82.5\scriptstyle\pm0.2$ & $83.2\scriptstyle\pm0.2$\tabularnewline
    & top-5 & $90.0\scriptstyle\pm0.1$ & $90.7\scriptstyle\pm0.16$ & $92.3\scriptstyle\pm0.3$ & $92.3$ &$--$&$--$& $94.1\scriptstyle\pm0.8$& $94.8\scriptstyle\pm0.2$&\tabularnewline
    \multirow{2}{*}{ILSVRC12} & top-1 & $72.1\scriptstyle\pm0.4$ & $74.4\scriptstyle\pm0.29$& $78.6\scriptstyle\pm0.4$ &$75.4$ &$--$&$--$& $79.9\scriptstyle\pm0.2$ & $80.6\scriptstyle\pm0.3$\tabularnewline
    & top-5 & $89.1\scriptstyle\pm0.3$ &$91.2\scriptstyle\pm0.12$ & $93.0\scriptstyle\pm0.1$ &$92.4$ &$--$&$--$& $94.6\scriptstyle\pm0.6$ & $95.2\scriptstyle\pm0.1$\tabularnewline
       \bottomrule
    \end{tabular}}
     
\end{table*}

\section{mini-Webvision results with ResNet50~\label{sec:webvisres50}}
We report in Table~\ref{tab:webvisres50} results for noise-robust algorithms training a ResNet50 on the mini-Webvision dataset. We report results for Contrast to Divide (C2D)~\cite{2022_WACV_C2D} that trains DivideMix (DM)~\cite{2020_ICLR_DivideMix} from a SimCLR initialization, Twin Contrastive Learning (TCL)~\cite{2023_CVPR_TCL} that trains a two-head contrastive network where the distribution of ID and OOD samples are captured by a two mode Gaussian mixture model, Label Confidence Incorporation (LCI)~\cite{2023_ICCV_LCI} that uses a teacher network trained on noisy data to supervise a noise-free student model and Neighbor Consistency Regularization (NCR)~\cite{2022_CVPR_NeighborConsistency} that regularizes samples close in the feature space to have similar supervised predictions. Similarly to results using InceptionResNetV2 in the main body of the paper, PLS-LSA improves over related work from $1$ to $2$ accuracy points and PLS-LSA+ further improves the results of PLS-LSA by $0.5$ to $1$ absolute point.

\section{Are co-training benefits only limited to network ensembling at test time ?\label{sec:cotraining}}

Because co-training is now a common strategy for label noise robustness as many newer methods~\cite{2023_PR_longremix,2023_ICCV_rankmatch,2023_PR_scanmix} build up on DivideMix (DM)~\cite{2020_ICLR_DivideMix}, we aim to find out if co-training strategies produces better individual networks or if they are better simply because a network ensemble is used at test time. \\
We train PLS-LSA+ using the following co-training strategies: an independent approach  (Indep) where the only interactions the two networks have is the test time prediction ensemble, the DivideMix co-training strategy (DM) where a network predicts noisy samples for the other and semi-supervised imputation is done using the ensemble prediction of the networks, a naive voting strategy (Vote) where the noisy samples are selected when detected as noisy by both networks (also ensembling for SSL imputation) and our co-training strategy (Ours) where we use the voting strategy but use a co-guessing strategy for the pseudo-loss of PLS (one network validates the SSL imputation for the other). \\
We report results training PLS-LSA+ using these co-training strategies for noise ratios $0.2$ and $0.8$ on the CNWL dataset in Table~\ref{tab:cotrain2}. The results are displayed as best accuracy for the ensemble (Ens) and for the individual (Indiv) networks. We additionally report the p-value obtained from a T-test of the current strategy against the independent one to evaluate if the improvement of the current co-train strategy are statistically better than the independent strategy. \\
We find that our co-training is the strategy producing the most statistically significantly more accurate individual networks ($p<0.05$) and that the semi-supervised co-validation strategy is important to acheive improved individual networks (Ours vs Vote). We recommend that future label noise research utilizing co-training strategies to conduct similar experiments to prove that the co-training strategy is beneficial beyond network ensembling.

\begin{table*}[t]
\centering{}\caption{Is co-training better than network ensembling ? We report the p-value of each strategy against the independent one. CNWL dataset. We bold the \textbf{best accuracy} and underline \underline{p-values under $0.05$}\label{tab:cotrain2}}
\global\long\def\arraystretch{0.9}%
\resizebox{\linewidth}{!}{{{}}

\begin{tabular}{l>{\centering}c>{\centering}c>{\centering}c>{\centering}c|>{\centering}c>{\centering}c>{\centering}c>{\centering}c>{\centering}c>{\centering}c>{\centering}c}
& \multicolumn{4}{c}{$20\%$ noise} & \multicolumn{4}{c}{$80\%$ noise} \tabularnewline
\cmidrule{2-5} \cmidrule{6-9}
& Indep & DM & Vote & Ours & Indep & DM & Vote & Ours  \tabularnewline
\midrule
Ens. & $66.15 \scriptstyle\pm 0.23$ & $66.24 \scriptstyle\pm 0.22$ &  $66.23 \scriptstyle\pm 0.18$ &  $\textbf{66.64} \scriptstyle\pm 0.37$ & $50.61 \scriptstyle\pm 0.49$ &  $50.83 \scriptstyle\pm 0.01$ & $50.76 \scriptstyle\pm 0.48$ &  $\textbf{51.88} \scriptstyle\pm 0.36$ \tabularnewline
\cmidrule{2-5}\cmidrule{6-9}
p-value & $1$ & $0.72$ & $0.74$ & $0.19$ & $1$ & $0.20$ & $0.85$ & $0.20$ \tabularnewline
\midrule
Indiv. & $63.95 \scriptstyle\pm 0.42$ & $64.56 \scriptstyle\pm 0.46$ &  $64.52 \scriptstyle\pm 0.13$ & $\textbf{64.71} \scriptstyle\pm 0.28$ & $47.50 \scriptstyle\pm 0.48$ & $48.34 \scriptstyle\pm 0.10$  & $48.38 \scriptstyle\pm 0.74$ & $\textbf{49.07} \scriptstyle\pm 0.31$ \tabularnewline
\cmidrule{2-5}\cmidrule{6-9}
p-value & $1$ & $0.06$ & $\underline{0.015}$ & $\underline{0.007}$ & $1$ & $\underline{0.05}$ & $0.28$ & $\underline{0.03}$ \tabularnewline
\bottomrule
\end{tabular}}
\end{table*}

\begin{figure*}[t]
\centering
\includegraphics[width=.9\linewidth]{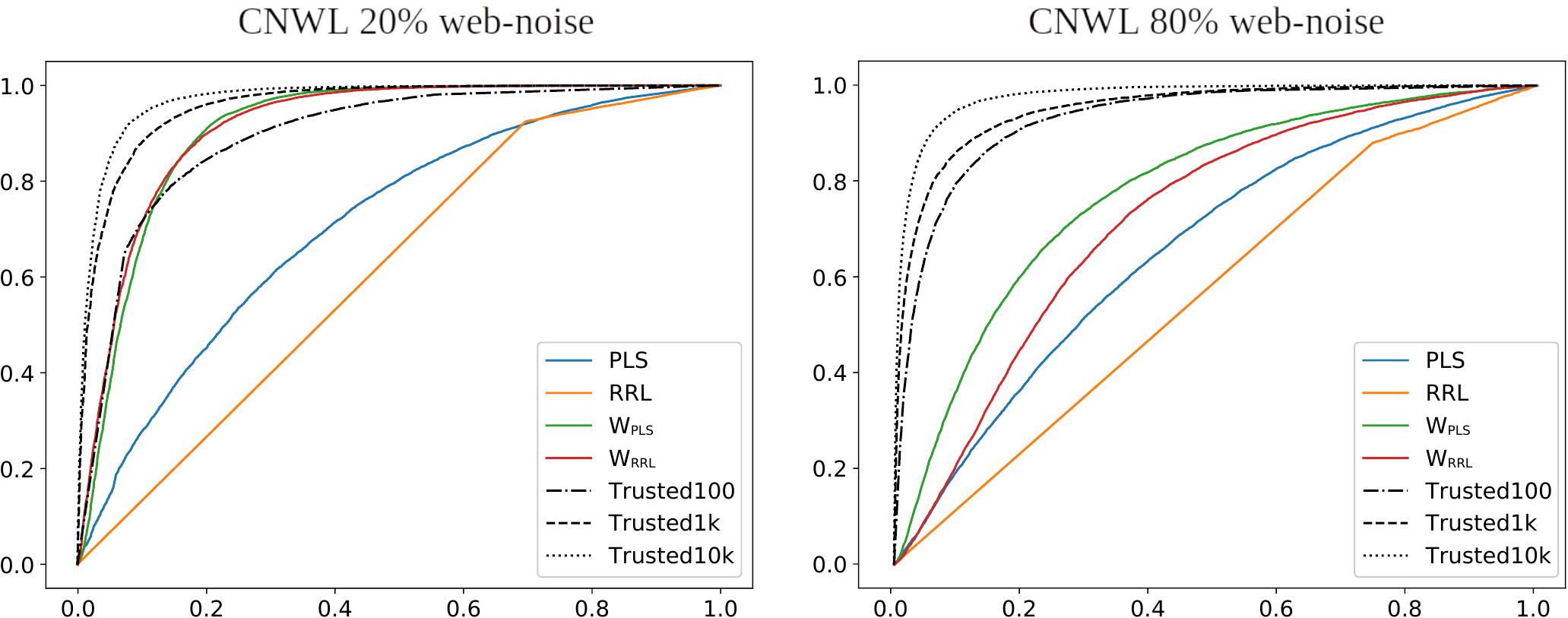}
\caption{ROC for different noise-retrieval metrics. We report PLS (loss-based) and RRL (feature-based), the refined detection when they are used as a support set for the logistic regressor ($W_{PLS}$ and $W_{RRL}$ respectively) and results where trusted examples (100, 1k or 10k) are used for training the logistic regressor. Features extracted after the block 2 of a PreAct ResNet18. \label{fig:roc}}
\end{figure*}

\section{Human labeled subset~\label{sec:stratslsa}}
\subsection{Improved noise retrieval using a trusted subset}
We visualize here the noise retreival capacities of PLS, RRL, $W_{PLS}$ and $W_{RRL}$ by plotting the Receiver Operating Characteristic Curves (ROC) when identifying noisy samples on the CNWL dataset under $20\%$  and $80\%$ noise. Although this is not a case we study in this paper, we additionally report here the performance of utilizing a human annotated subset (oracle) to compute $W$. We run experiments where we train the logistic regressor on $10.000$ (10k), $1.000$ (1k) and $100$ randomly selected and ID/OOD-annotated samples.  Figure~\ref{fig:roc} reports the results. We find that $W_{PLS}$ and $W_{RRL}$ import on the metrics that are based on. As little as $100$ trusted samples provide a strong noise detection especially in high noise ratios. We leave this information for future work.

\begin{table}[t]
\centering{}\caption{Training PLS-LSA using trusted subsets. CNWL dataset with various noise levels. PLS-LSA uses $W_{PLS}$ for the linear separation while others use a trusted subset e.g $W_{100}$. Results averaged over 3 runs $\pm$ one std~\label{tab:trusted}}

\global\long\def\arraystretch{0.9}%
\resizebox{.7\linewidth}{!}{{{}}%
\begin{tabular}{l>{\centering}c>{\centering}c>{\centering}c>{\centering}c>{\centering}c}
\toprule
Noise ratio & $0.2$ & $0.4$ & $0.6$ & $0.8$ \tabularnewline
\midrule
PLS-LSA & $64.43 \scriptstyle\pm 0.21$ & $61.14 \scriptstyle\pm 0.35$ & $57.18 \scriptstyle\pm 0.30$ & $49.53 \scriptstyle\pm 0.46$ \tabularnewline
100 & $64.51 \scriptstyle\pm 0.34$ & $61.48 \scriptstyle\pm 0.20$ & $58.07 \scriptstyle\pm 0.33$ & $50.35 \scriptstyle\pm 0.43$ \tabularnewline
1k & $64.62 \scriptstyle\pm 0.31$ & $61.83 \scriptstyle\pm 0.18$ & $58.16 \scriptstyle\pm 0.55$ & $50.84 \scriptstyle\pm 0.38$ \tabularnewline
10k & $64.88 \scriptstyle\pm 0.31$ & $62.07 \scriptstyle\pm 0.30$ & $58.61 \scriptstyle\pm 0.25$ & $51.43 \scriptstyle\pm 0.40$ \tabularnewline
\iffalse
\midrule
\multicolumn{5}{c}{Co-train} \tabularnewline
\midrule
PLS-LSA+ & $66.52 \scriptstyle\pm 0.10$ & $63.42\scriptstyle\pm 0.42$ & $59.41 \scriptstyle\pm 0.30$ & $52.03 \scriptstyle\pm 0.32$ \tabularnewline
100 & $66.37 \scriptstyle\pm 0.19$ & $63.69 \scriptstyle\pm 0.23$ & $59.64 \scriptstyle\pm 0.17$ & $52.99 \scriptstyle\pm 0.09$ \tabularnewline
1k & $67.21 \scriptstyle\pm 0.25$ & $63.95 \scriptstyle\pm 0.23$ & $59.61 \scriptstyle\pm 0.20$ & $52.95 \scriptstyle\pm 0.13$ \tabularnewline
10k & $67.59 \scriptstyle\pm 0.12$ &  $64.24 \scriptstyle\pm 0.09$ & $60.03 \scriptstyle\pm 0.09$ & $53.43 \scriptstyle\pm 0.14$ \tabularnewline
\fi
\bottomrule
\end{tabular}}
\end{table}

\subsection{PLS-LSA with a trusted subset~\label{sec:trusted}}
We utlize here the trusted subset computed in the previous subsection to train PLS-LSA/ Results and the comparison against our unsupervised solution can be found in Table~\ref{tab:trusted}. We observe that for up to $40\%$ noise corruption, our unsupervised approach performs on par with using $100$ to $1.000$ trusted samples yet the added supervision of even $100$ trusted ID/OOD samples becomes beneficial for the $60\%$ and $80\%$ noise scenarios.

\subsection{Different strategies to alternate feature and loss detection\label{sec:alternstrat}}
We study here different strategies for alternating between $Z$ and $W$. We propose to compare the approach proposed in the main body of the paper (modulo 2) against a random choice every epoch with a probability of $50\%$ (random) or a random choice for each training sample (random sample) at a given epoch instead of using the same strategy for all samples. Results are displayed for the CNWL dataset under noise perturbations of $0.2$ and $0.8$ in Table~\ref{tab:randselection}. We observe that alternating randomly between $Z$ and $W$ is similarly accurate than regulated alternation every other epoch (modulo 2). Doing a random selection at the sample level (random sample) is however less accurate. These results appear to evidence that maintaining a selection logic (linear separation or small-loss) for a period of time of at least one epoch in our case is beneficial.

\begin{table}[t]
\centering{}\caption{Different alternating strategies for LSA~\label{tab:randselection}}
\global\long\def\arraystretch{0.9}%
\resizebox{0.5\linewidth}{!}{{{}}%
\begin{tabular}{l>{\centering}c>{\centering}c>{\centering}c>{\centering}c}
\toprule
 Noise ratio & 0.2 & 0.8 \tabularnewline
 \midrule
modulo 2 & $64.43 \scriptstyle\pm 0.21$ & $49.53 \scriptstyle\pm 0.46$ \tabularnewline
random & $64.44 \scriptstyle\pm 0.02$ & $48.32 \scriptstyle\pm 0.17$ \tabularnewline
random sample & $63.15 \scriptstyle\pm 0.24$ & $46.21 \scriptstyle\pm 0.58$ \tabularnewline
 \bottomrule
\end{tabular}}
\end{table}

\begin{table}[t]
\centering{}\caption{Using features after different ResNet blocks to compute $W_{PLS}$, CNWL.~\label{tab:diffl}}

\global\long\def\arraystretch{0.9}%
\resizebox{.6\linewidth}{!}{{{}}%
\begin{tabular}{l>{\centering}c>{\centering}c>{\centering}c>{\centering}c>{\centering}c}
\toprule
  Noise & $0.2$ & $0.4$ & $0.6$ & $0.8$ \tabularnewline
 \midrule
 0 & $64.19 \scriptstyle\pm 0.38$ & $60.66 \scriptstyle\pm 0.41$ & $57.00 \scriptstyle\pm 0.15$ & $48.41 \scriptstyle\pm 0.10$ \tabularnewline
 1 & $64.43 \scriptstyle\pm 0.21$ & $61.14 \scriptstyle\pm 0.35$ & $57.18 \scriptstyle\pm 0.30$ & $49.53 \scriptstyle\pm 0.46$ \tabularnewline
 2 & $63.79 \scriptstyle\pm 0.27$ & $60.35 \scriptstyle\pm 0.29$ & $56.92 \scriptstyle\pm 0.16$ & $49.25 \scriptstyle\pm 0.62$ \tabularnewline
 3 & $63.41 \scriptstyle\pm 0.36$ & $60.19 \scriptstyle\pm 0.14$ & $56.52 \scriptstyle\pm 0.27$ & $49.01 \scriptstyle\pm 0.29$ \tabularnewline
 4 & $63.73 \scriptstyle\pm 0.03$ & $59.92 \scriptstyle\pm 0.03$ & $55.79 \scriptstyle\pm 0.17$ & $48.45 \scriptstyle\pm 0.78$ \tabularnewline
 \bottomrule
\end{tabular}}
\end{table}

\subsection{Computing the linear separation at different depth\label{sec:linsepdepth}}
We study here the influence of computing the linear separation on features at different depth in the network. We run PLS-LSA utilizing features extracted after blocks 0-3 in the ResNet18 architecture as well as utilizing the contrastive projection (block 4 in this case). The results can be found in Table~\ref{tab:diffl}. We find that features extracted at block 1 produce the more accurate networks and that the accuracy degrades when using deeper layers. These results are coherent with the observed noise retrieval accuracy using the linear separation in the main body of the paper. For every experiment where we run PLS-LSA, we use average-pooled then L2 normalized features at the end of the 2nd block of our ResNet architecture (feature dimension $128$) for $W$.

\section{Missed important samples\label{sec:missedimp}}

\iffalse
\subsection{No class left behind}
To show that the lower than expected validation accuracy obtained when training our linear separation in Table~\ref{tab:strongercleannoisy} is not due to some classes beeing missed, we show in Figure~\ref{fig:top20worse} the AUROC for the 20 classes of the CNWL under $20\%$ noise where our linear separation (10k trusted samples in this case) performs the worse. We also report results for the noise retrieval performance of the PLS metric on the same classes. We find that worse class bound AUROC for the linear separation is $78.2$ which is far from a random guess of $50$. This comforts us in the fact that the low performance of our accurate detection is not due to some classes being wrongly detected.

\begin{figure}
    \centering
    \includegraphics[width=.8\linewidth]{images/LowAccurayTrusted10kvsPLS-crop.pdf}
    \caption{Top 20 worse performing classes for our linear separation detection under 10k trusted samples. CNWL $20\%$ noise. \label{fig:top20worse}}
\end{figure}
\fi
\subsection{Complementary noise detections\label{sec:complmissed}}

\begin{figure}[t]
    \centering
    \includegraphics[width=.6\linewidth]{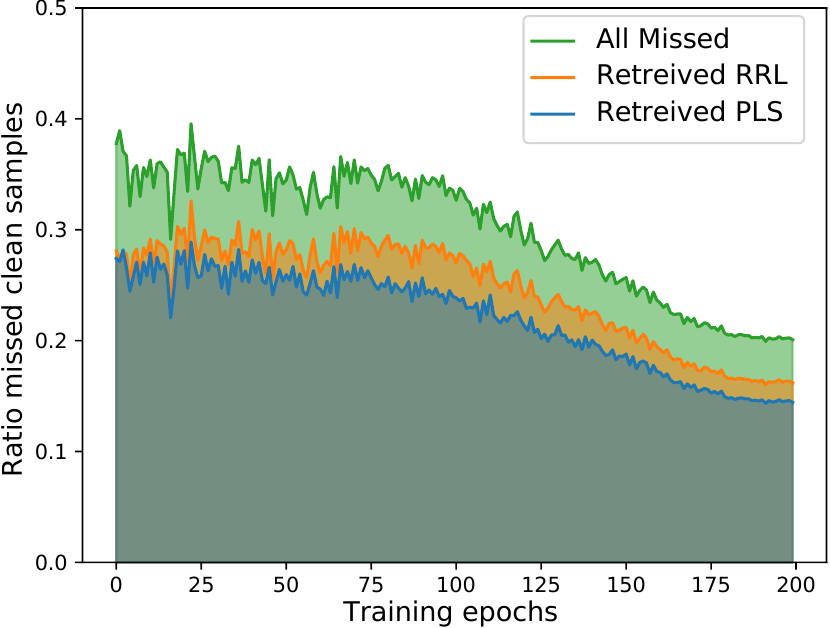}
    \caption{Clean samples missed by our linear separation but retrieved by PLS or RRL. PLS-LSA trained on the CNWL $20\%$. \label{fig:compl}}
\end{figure}

We report how complementary our linear separation retrieval is with a generic small loss approach by plotting every epoch of training PLS-LSA how much of the missed clean samples is retrieved by either PLS or RRL. Figure~\ref{fig:compl} displays the results where we observe that up to $80\%$ of the missed samples are retrieved and that the further the PLS-LSA training progresses, the less clean samples our linear separation misses (from $40\%$ of the total clean samples at the start of training to less than $20\%$ at the end).

\begin{figure}[t]
    \centering
    \includegraphics[width=.7\linewidth]{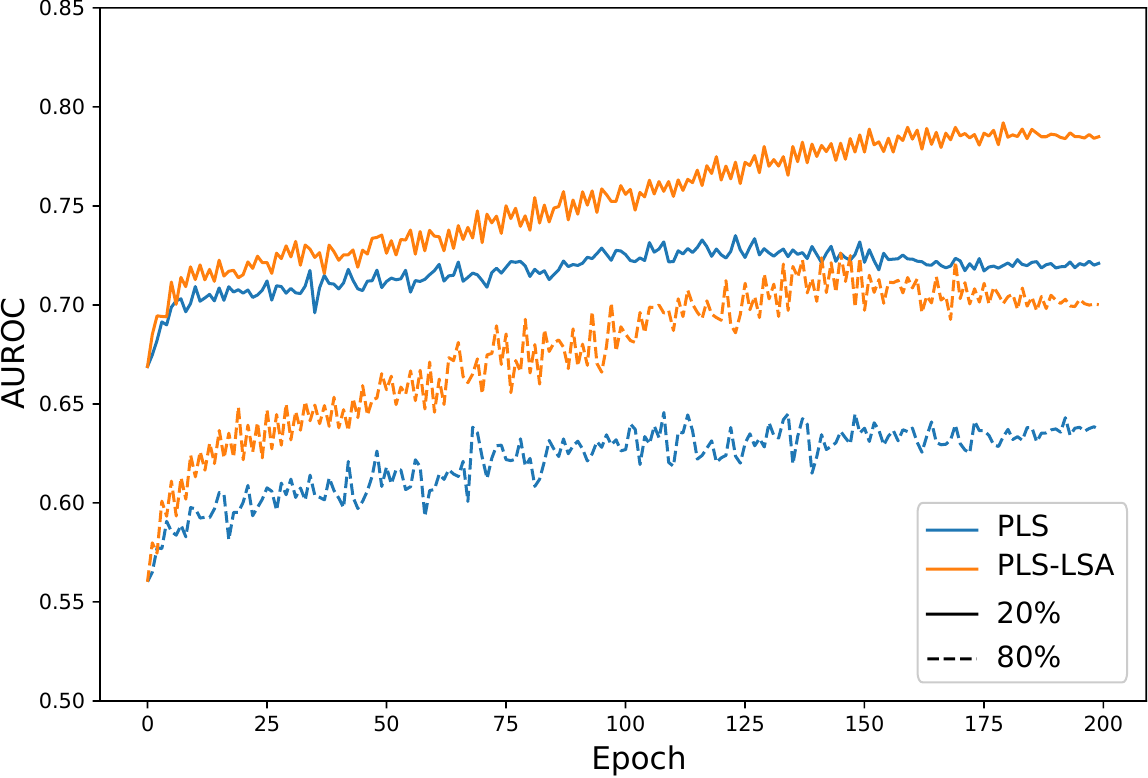}
    \caption{PLS-LSA improves the small loss noise retrieval of PLS. CNWL under $20\%$ or $80\%$ noise. \label{fig:improvedret}}
\end{figure}

\subsection{LSA improves small loss noise detection\label{sec:improving}}
Another observation we make of the mutual benefits of our linear separation alternating is the improved noise retrieval of the original PLS metric (small loss) when training PLS-LSA as opposed to PLS alone. Figure~\ref{fig:improvedret} reports the AUROC for the PLS noise detection metric retrieving noisy samples in the PLS or PLS-LSA configurations. We observe that the small loss retrieval of PLS is improved when trained with LSA, highlight the complementarity and resulting mutual improvement of each metric.
\iffalse
\begin{table*}[]
\centering{}\caption{Influence of class-balanced regularization. Averaged over 3 random seeds $\pm$ std~\label{tab:ablapen}}
\global\long\def\arraystretch{0.9}%
\resizebox{.6\linewidth}{!}{{{}}%
\begin{tabular}{l>{\centering}c>{\centering}c>{\centering}c>{\centering}c>{\centering}c>{\centering}c>{\centering}c>{\centering}c>{\centering}c>{\centering}c>{\centering}c}
\toprule
  & \multicolumn{2}{c}{$20\%$} & \multicolumn{2}{c}{$80\%$} \tabularnewline
 \cmidrule(lr){2-3} \cmidrule(lr){4-5}
   & PLS & PLS-LSA & PLS & PLS-LSA \tabularnewline
 \midrule 
 no penalty &  $63.71 \pm 0.13$ & $64.55 \pm 0.34$ &  $48.64 \pm 0.16$ & $48.91 \pm 0.30$ \tabularnewline
 PLS pen & $63.44 \pm 0.12$ & $64.54 \pm 0.52$ & $48.68 \pm 0.13$ & $49.49 \pm 0.38$ \tabularnewline
 our pen & $63.62 \pm 0.10$ & $64.61 \pm 0.51$ & $47.98 \pm 0.44$ & $48.20 \pm 0.16$ \tabularnewline
 both pen & $64.05 \pm 0.07$ & $64.51 \pm 0.14$ & $47.53 \pm 0.34$ & $48.91 \pm 0.20$ \tabularnewline
 \bottomrule
\end{tabular}}
\end{table*}
\fi

\begin{figure}
    \centering
    \includegraphics[width=\linewidth]{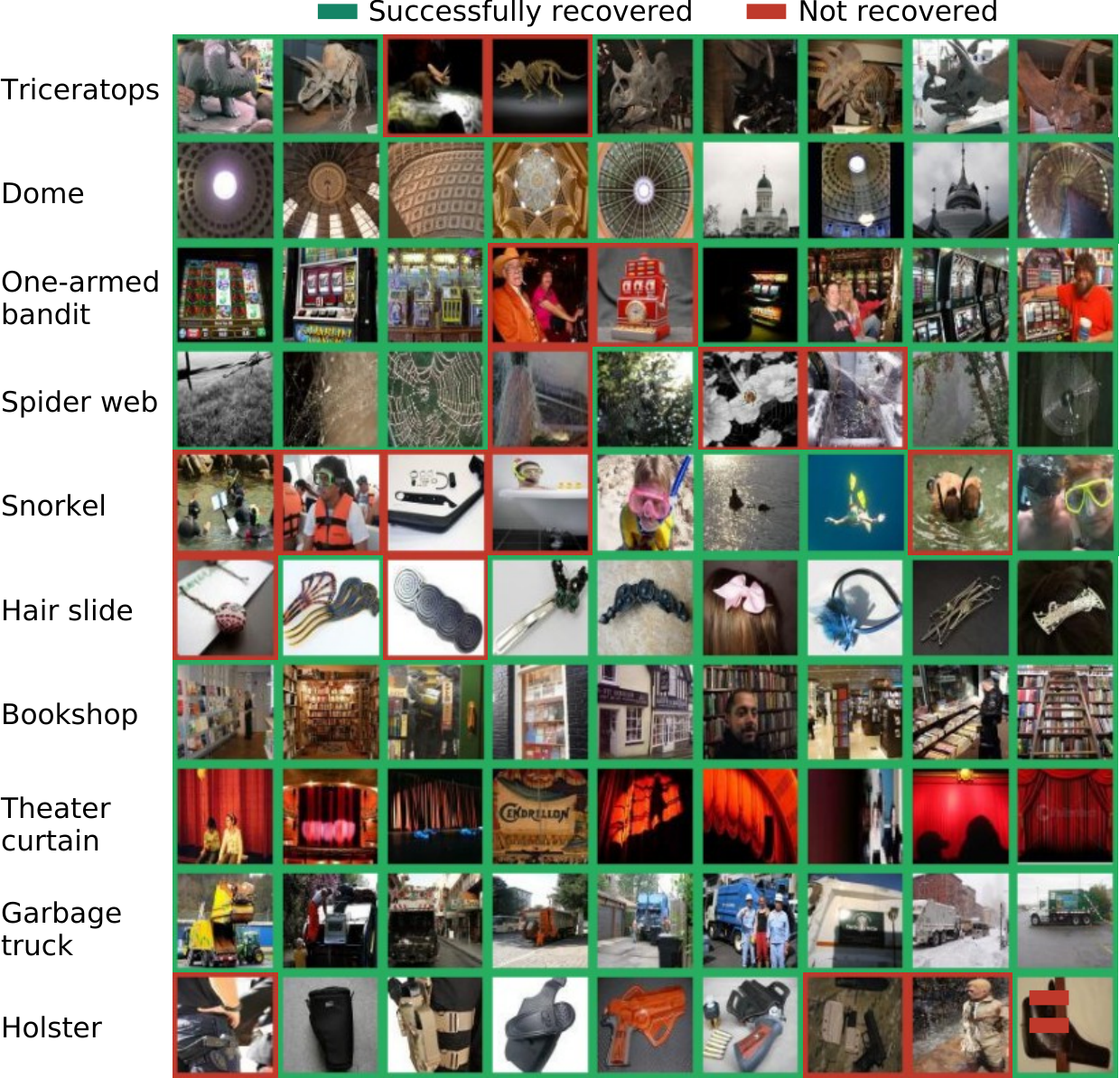}
    \caption{Examples of clean samples missed by our linear separation $W_{PLS}$ but correctly recovered by PLS (green). $20\%$ noise CNWL. Repeated from the main body for convenience\label{fig:missedLin}}
\end{figure}

\begin{figure}
    \centering
    \includegraphics[width=\linewidth]{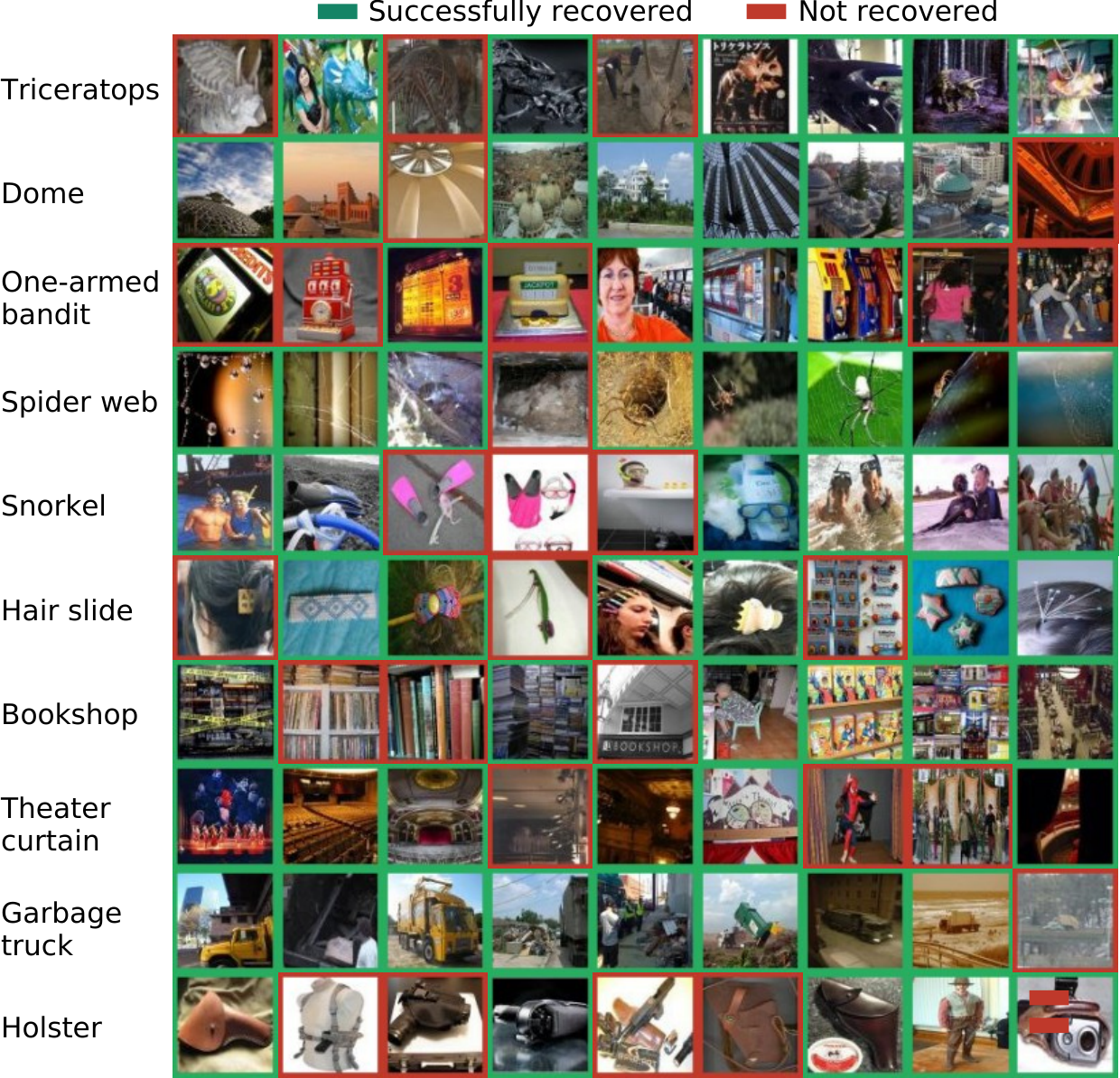}
    \caption{Examples of clean samples missed by PLS but correctly recovered by our linear separation $W_{PLS}$ (green). $20\%$ noise CNWL.\label{fig:missedPLS}}
\end{figure}
\subsection{Visualizing missed clean samples\label{sec:missedsamples}}
Figure~\ref{fig:missedLin} displays examples of clean samples missed by our linear separation detection.  We display images for classes 0, 15, 17, 25, 36, 45, 59, 61, 95 and 96 of the CNWL dataset (randomly selected). We notice how most of the missed samples are the target object displayed on a uniform background free of distractors. We also report in Figure~\ref{fig:missedPLS} the opposite scenario: clean images missed by PLS but successfully identified as clean by our linear separation. In this second scenario, we observe that the images missed by PLS but retrieved by our linear separation appear to be more difficult images with a cluttered background or presenting a small instance of the target class.

\section{Additional results for CLIP ViT architectures and other noise robust algorithms}
\subsection{CLIP architectures}
We provide here some results on training PLS-LSA on ViT architectures pre-trained using a CLIP-like framework~\cite{2021_ICML_CLIP}. We obtain pretrained weights from the open-clip repository~\cite{2023_CVPR_openclip} and finetune the ResNet-50 and ViT-B/32 architectures on the CNWL and Webivison datasets. Results are reported in Table~\ref{tab:clip} where we add non-robust training with mixup and PLS as baselines. We find that LSA scales well when applied to the CNWL dataset but Webvision improvements are less convincing, supposedly because the margin for improvement is small. These early results suggest that LSA is generalizable to transformer architectures and different manners of contrastive pre-training.

\begin{table}[t]
    \vspace{-10pt}  
    \centering
    \caption{PLS-LSA with CLIP. Top-1 accuracy}
    \label{tab:clip}
    \resizebox{.6\linewidth}{!}{{{}}
    \begin{tabular}{l>{\centering}c>{\centering}c>{\centering}c>{\centering}c>{\centering}c}
    \toprule
    & Dataset & CNWL 0.4 & CNWL 0.8 & Webvision \tabularnewline
    \midrule
    \multirow{3}{*}{ViT-B/32} & mixup & 84.06 & 78.94 & 83.80 \tabularnewline
    & PLS & 85.72 & 80.04 & 85.12 \tabularnewline
    & PLS-LSA & 86.70 & 82.22 & 85.20 \tabularnewline
    \midrule
    \multirow{3}{*}{ResNet-50} & mixup & 77.54 & 69.56 & 81.88 \tabularnewline
    & PLS & 77.72 & 70.04 & 82.36 \tabularnewline
    & PLS-LSA & 78.60 & 71.78 & 82.64 \tabularnewline
    \bottomrule
    \end{tabular}}
\end{table}

\subsection{ProMix-LSA}
We report here additional results when training LSA with ProMix~\cite{2023_IJCAI_promix} the current leader of the CIFAR-N datasets~\cite{2022_ICLR_CIFARN} leaderboard~\footnote{http://noisylabels.com/}. We compare ProMix-LSA with PLS-LSA+ as ProMix utilizes an ensemble of two networks to predict. We report results on the CNWL dataset and Webvision at the resolution of $32\times32$ as ProMix requires an amount of VRAM too large to train at full resolution with our resources. ProMix-LSA largely improves over ProMix alone in all scenarios. 

\begin{table}[t]
    \centering   
    \caption{LSA applied to ProMix. Top-1 accuracy with a PreActivation ResNet18.}
    \label{tab:promix}
    \global\long\def\arraystretch{0.9}%
    \resizebox{.6\linewidth}{!}{{{}}
    \begin{tabular}{l>{\centering}c>{\centering}c>{\centering}c>{\centering}c>{\centering}c}
    \toprule
    Dataset & CNWL 0.4 & CNWL 0.8 & Webvision (32x32)\tabularnewline
    \toprule
    ProMix & 60.70 & 46.54 & 64.80 \tabularnewline
    ProMix-LSA & 64.00 & 50.96 & 68.12     \tabularnewline
    \midrule
    PLS-LSA+ & 63.42 & 52.03 & 69.16 \tabularnewline
    \bottomrule
    \end{tabular}} 
    \vspace{-15pt}
\end{table}

\section{Top-n accuracy web-fg datasets\label{sec:topnwebfg}}
We report Top-2 and Top-5 accuracy results of PLS-LSA on the Web-fg datasets in Table~\ref{tab:sotawebfgtop5}. We observe that top-2 accuracy offers a significantly improvement over top-1 classification which indicates that PLS-LSA rarely catastrophically fails as if the target class is not the most accurate prediction is is often the second best.

\begin{table}[t]
    \centering
    \caption{Top-K classification accuracy of PLS-LSA on the Webly-fg datasets\label{tab:sotawebfgtop5}}
    \global\long\def\arraystretch{0.8}%
    \resizebox{.5\linewidth}{!}{{{}}%
    \begin{tabular}{l>{\centering}c>{\centering}c>{\centering}c>{\centering}c>{\centering}c}
    \toprule
        & Web-Aircraft & Web-bird & Web-car \tabularnewline
       \midrule
       Top-1 & $87.82$ & $79.47$ & $86.76$ \tabularnewline
       Top-2 & $95.11$ & $84.73$ & $94.03$ \tabularnewline
       Top-5 & $97.54$ & $93.92$ & $97.57$ \tabularnewline
    \bottomrule
    \end{tabular}}
\end{table}

\section{Example of detected noisy samples\label{sec:noisyimages}}
\begin{figure}[t]
    \centering
    \includegraphics[width=\linewidth]{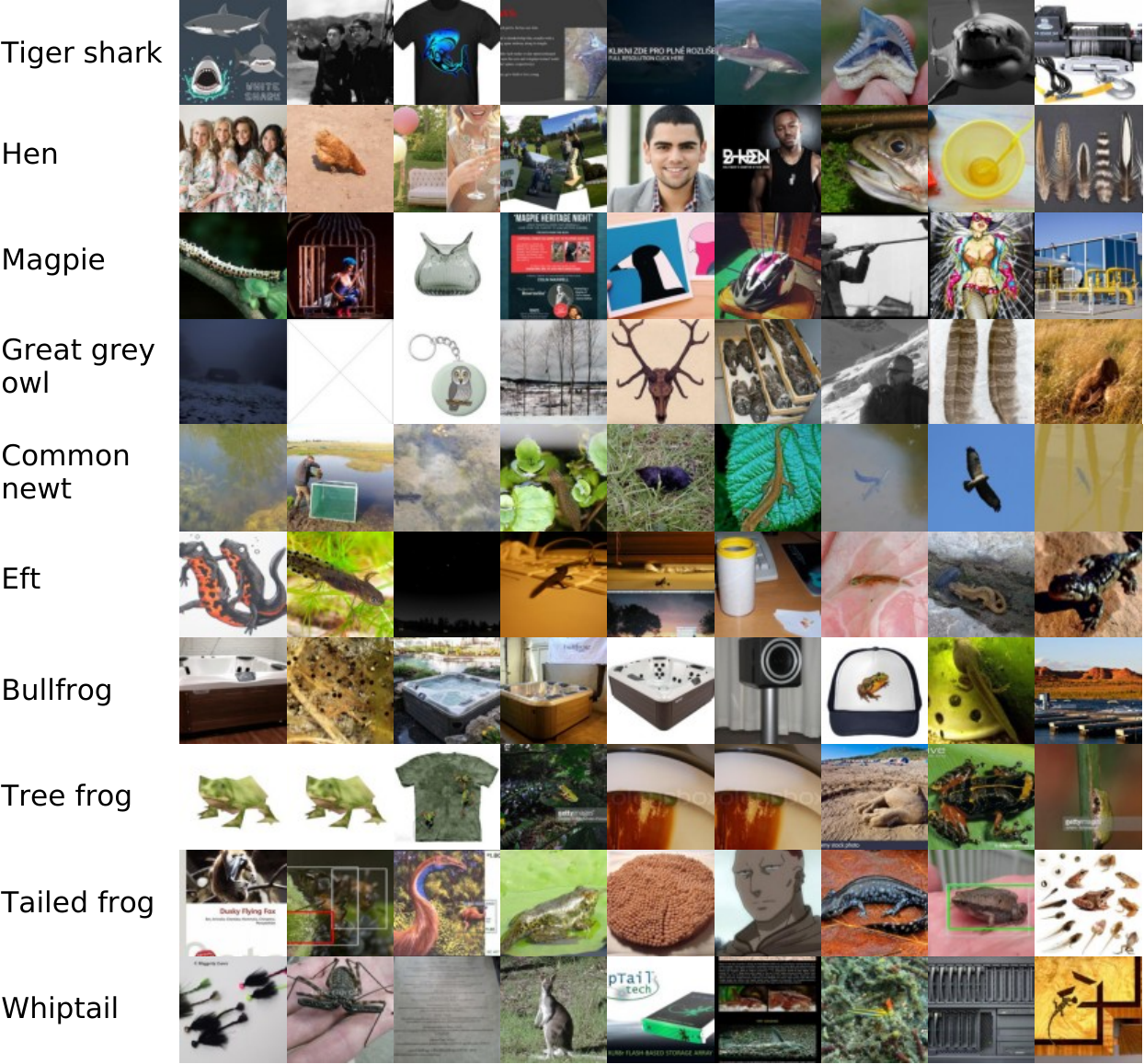}
    \caption{Examples of detected noisy samples in Webvision\label{fig:noisywebvis}}
\end{figure}

\begin{figure}[t]
    \centering
    \includegraphics[width=\linewidth]{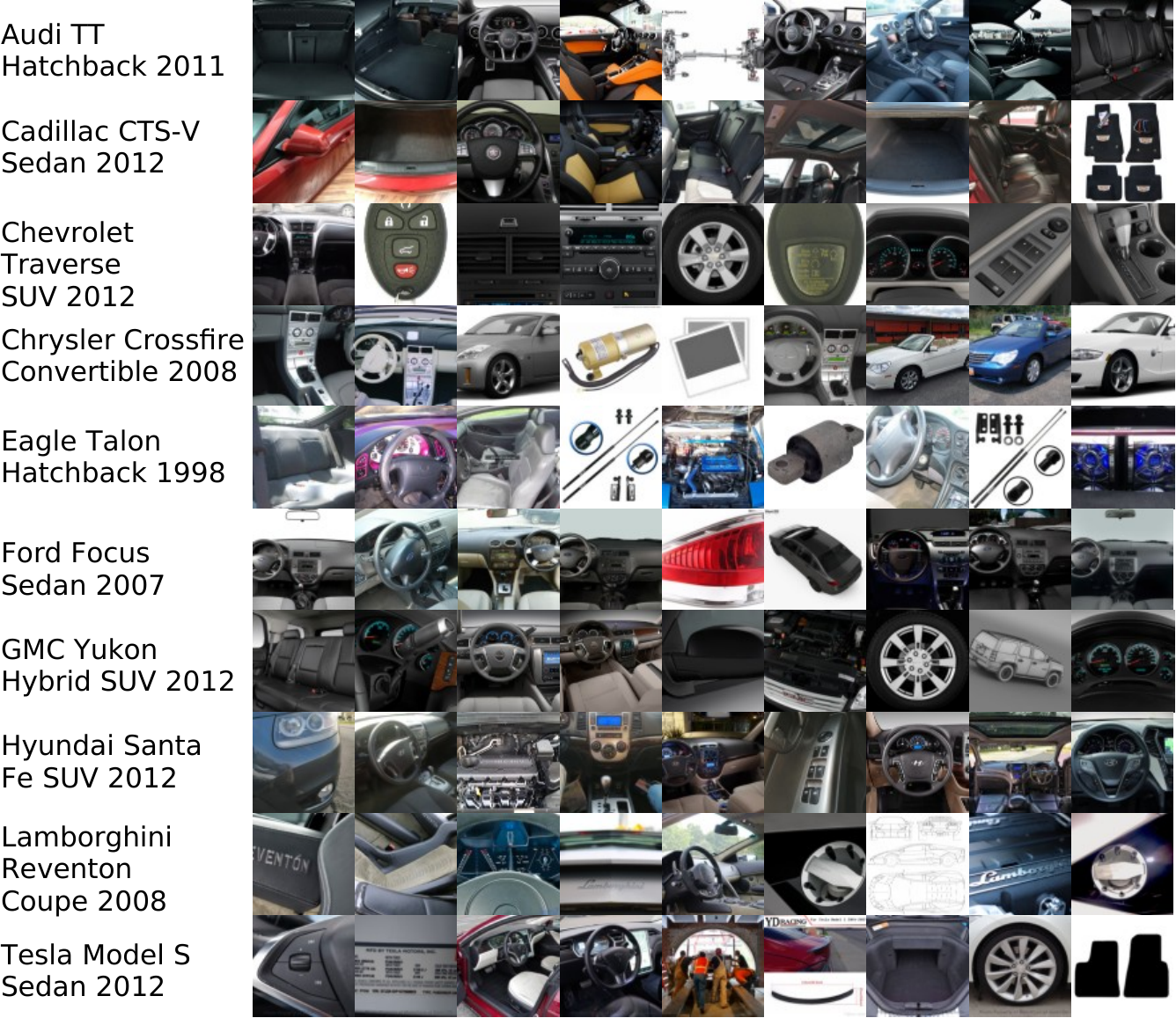}
    \caption{Examples of detected noisy samples in Web-car\label{fig:noisywebcar}}
\end{figure}

\begin{figure}[t]
    \centering
    \includegraphics[width=\linewidth]{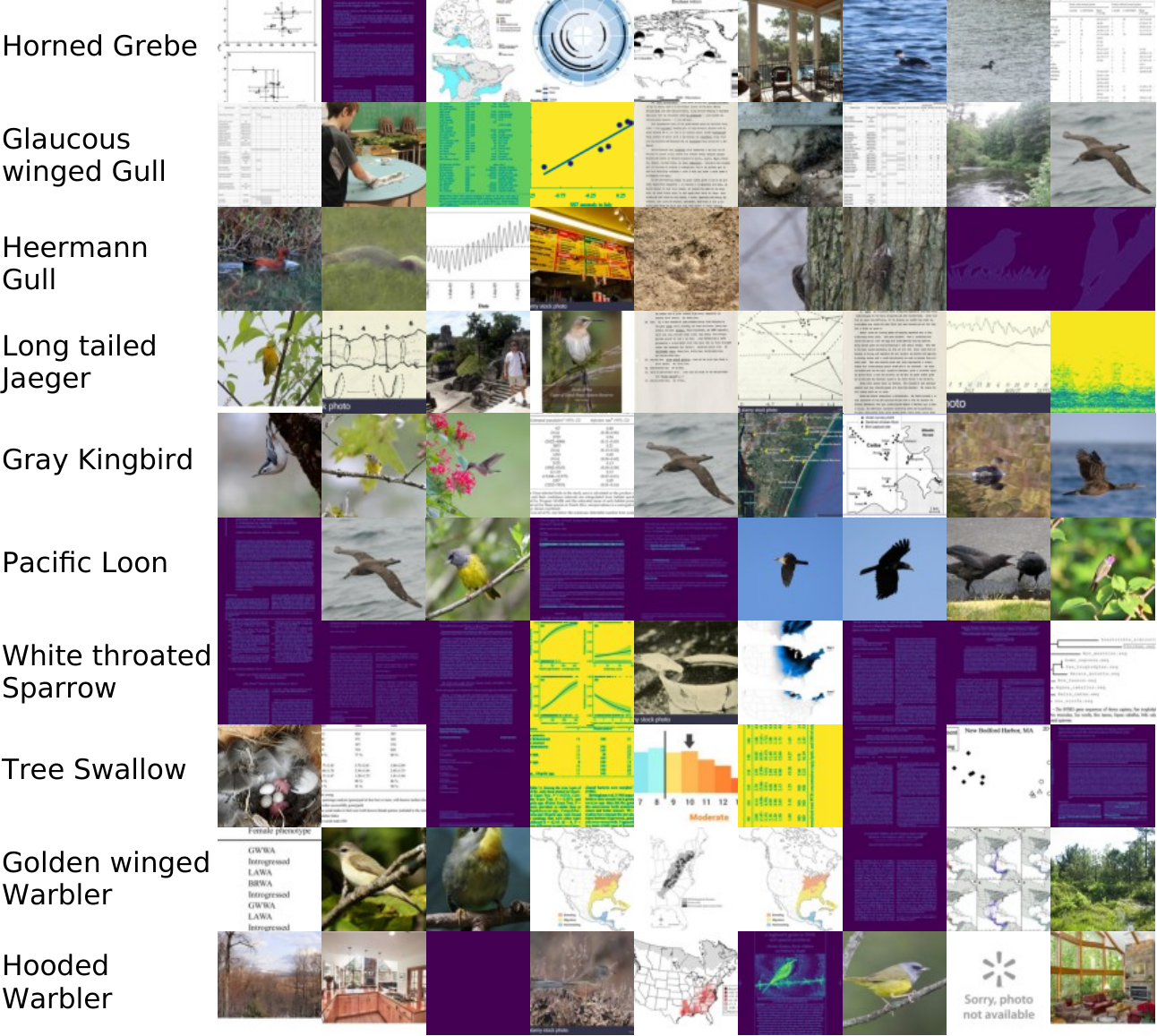}
    \caption{Examples of detected noisy samples in Web-bird\label{fig:noisywebbird}}
\end{figure}

\begin{figure}[t]
    \centering
    \includegraphics[width=\linewidth]{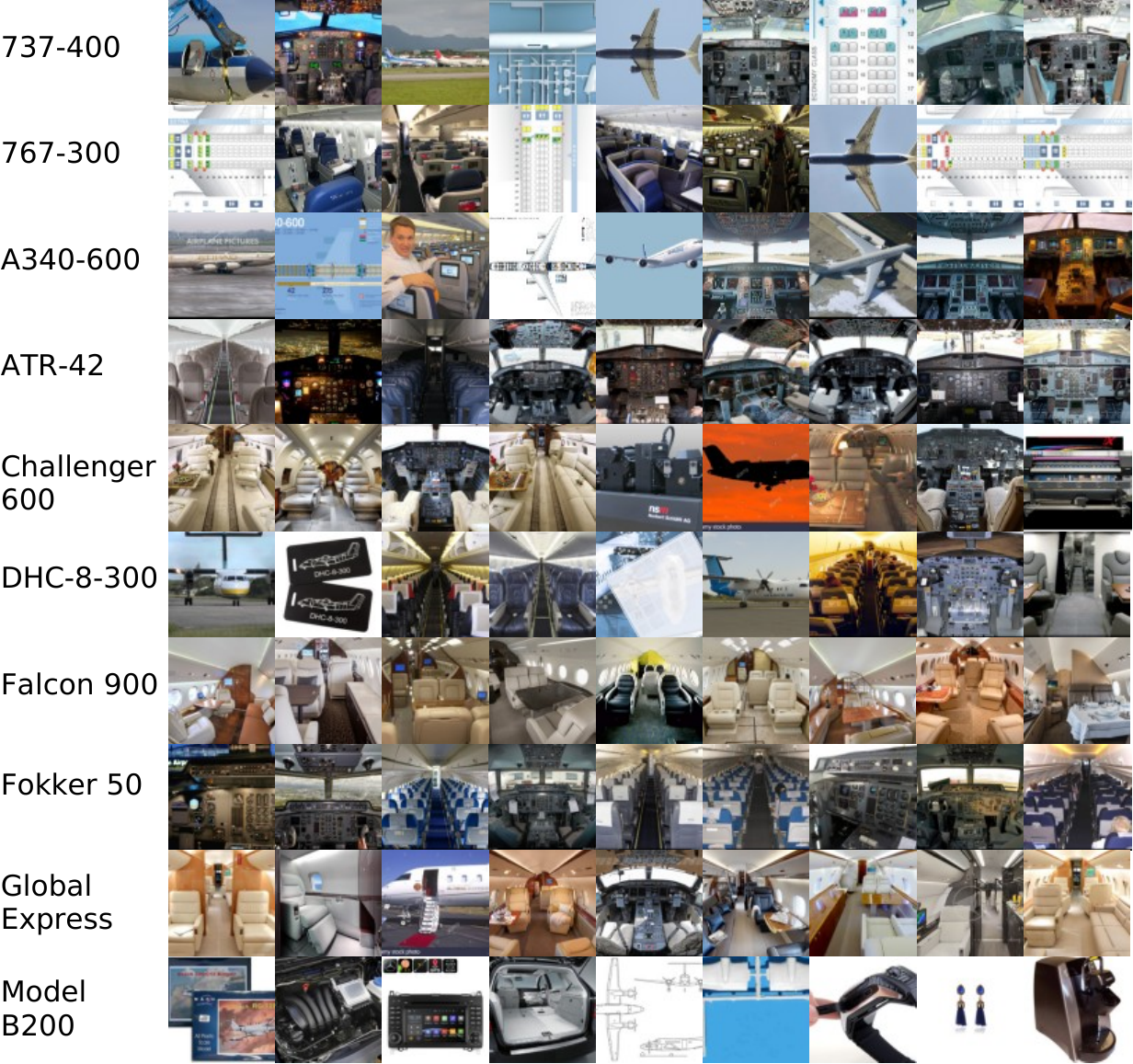}
    \caption{Examples of detected noisy samples in Web-aircraft\label{fig:noisywebaircraft}}
\end{figure}

Figure~\ref{fig:noisywebvis} reports examples of training samples we detect as noisy with PLS-LSA and Figures~\ref{fig:noisywebcar}, \ref{fig:noisywebbird} and \ref{fig:noisywebaircraft} report detected noisy examples on Web-{car/bird/aircraft}.

%% file: main.bbl
\begin{thebibliography}{10}
\providecommand{\url}[1]{\texttt{#1}}
\providecommand{\urlprefix}{URL }
\providecommand{\doi}[1]{https://doi.org/#1}

\bibitem{2023_ICCV_LCI}
Ahn, C., Kim, K., Baek, J.w., Lim, J., Han, S.: {Sample-wise Label Confidence Incorporation for Learning with Noisy Labels}. In: {Proceedings of the IEEE/CVF International Conference on Computer Vision} (2023)

\bibitem{2023_WACV_PLS}
Albert, P., Arazo, E., Krishna, T., O’Connor, N.E., McGuinness, K.: {Is your noise correction noisy? PLS: Robustness to label noise with two stage detection}. In: {IEEE/CVF Winter Conference on Applications of Computer Vision (WACV)} (2023)

\bibitem{2022_ECCV_SNCF}
Albert, P., Arazo, E., O’Connor, N.E., McGuinness, K.: {Embedding contrastive unsupervised features to cluster in-and out-of-distribution noise in corrupted image datasets}. In: {European Conference on Computer Vision (ECCV)} (2022)

\bibitem{2022_WACV_DSOS}
Albert, P., Ortego, D., Arazo, E., O'Connor, N., McGuinness, K.: {Addressing out-of-distribution label noise in webly-labelled data}. In: {Winter Conference on Applications of Computer Vision (WACV)} (2022)

\bibitem{1999_ACM_Optics}
Ankerst, M., Breunig, M.M., Kriegel, H.P., Sander, J.: Optics: Ordering points to identify the clustering structure. ACM Sigmod record  \textbf{28}(2),  49--60 (1999)

\bibitem{2019_ICML_BynamicBootstrapping}
Arazo, E., Ortego, D., Albert, P., O'Connor, N., McGuinness, K.: {Unsupervised Label Noise Modeling and Loss Correction}. In: {International Conference on Machine Learning (ICML)} (2019)

\bibitem{2020_IJCNN_Pseudo}
Arazo, E., Ortego, D., Albert, P., O'Connor, N., McGuinness, K.: {Pseudo-Labeling and Confirmation Bias in Deep Semi-Supervised Learning}. In: {International Joint Conference on Neural Networks (IJCNN)} (2020)

\bibitem{2019_NIPS_MixMatch}
Berthelot, D., Carlini, N., Goodfellow, I., Papernot, N., Oliver, A., Raffel, C.: {MixMatch: {A} Holistic Approach to Semi-Supervised Learning}. In: {Advances in Neural Information Processing Systems (NeuRIPS)} (2019)

\bibitem{2020_ICML_SimCLR}
Chen, T., Kornblith, S., Norouzi, M., Hinton, G.: {A Simple Framework for Contrastive Learning of Visual Representations}. In: {International Conference on Machine Learning (ICML)} (2020)

\bibitem{2023_CVPR_openclip}
Cherti, M., Beaumont, R., Wightman, R., Wortsman, M., Ilharco, G., Gordon, C., Schuhmann, C., Schmidt, L., Jitsev, J.: {Reproducible scaling laws for contrastive language-image learning}. In: {IEEE/CVF Conference on Computer Vision and Pattern Recognition (CVPR)} (2023)

\bibitem{2017_arXiv_ImageNet32}
Chrabaszcz, P., Loshchilov, I., Hutter, F.: {A downsampled variant of imagenet as an alternative to the cifar datasets}. arXiv: 1707.08819  (2017)

\bibitem{2021_arXiv_propmix}
Cordeiro, F.R., Belagiannis, V., Reid, I., Carneiro, G.: {PropMix: Hard Sample Filtering and Proportional MixUp for Learning with Noisy Labels}. arXiv: 2110.11809  (2021)

\bibitem{2023_PR_longremix}
Cordeiro, F.R., Sachdeva, R., Belagiannis, V., Reid, I., Carneiro, G.: {Longremix: Robust learning with high confidence samples in a noisy label environment}. {Pattern Recognition}  (2023)

\bibitem{2020_CVPRW_randaugment}
Cubuk, E.D., Zoph, B., Shlens, J., Le, Q.V.: {Randaugment: Practical automated data augmentation with a reduced search space}. In: {IEEE/CVF Conference on Computer Vision and Pattern Recognition Workshops (CVPRW)} ({2020})

\bibitem{2023_arXiv_MDM}
Fooladgar, F., To, M.N.N., Mousavi, P., Abolmaesumi, P.: {Manifold DivideMix: A Semi-Supervised Contrastive Learning Framework for Severe Label Noise}. arXiv:2308.06861  (2023)

\bibitem{2018_NeurIPS_CoTeaching}
Han, B., Yao, Q., Yu, X., Niu, G., Xu, M., Hu, W., Tsang, I., Sugiyama, M.: {Co-teaching: Robust training of deep neural networks with extremely noisy labels}. In: {Advances in Neural Information Processing Systems (NeurIPS)} (2018)

\bibitem{2016_ECCV_preactresnet}
He, K., Zhang, X., Ren, S., Sun, J.: {Identity mappings in deep residual networks}. In: {European Conference on Computer Vision (ECCV)} (2016)

\bibitem{2023_CVPR_TCL}
Huang, Z., Zhang, J., Shan, H.: {Twin contrastive learning with noisy labels}. In: {IEEE/CVF Conference on Computer Vision and Pattern Recognition (CVPR)} (2023)

\bibitem{2022_CVPR_NeighborConsistency}
Iscen, A., Valmadre, J., Arnab, A., Schmid, C.: {Learning With Neighbor Consistency for Noisy Labels}. In: {Proceedings of the IEEE/CVF Conference on Computer Vision and Pattern Recognition (CVPR)} (2022)

\bibitem{2018_ICML_MentorNet}
Jiang, L., Zhou, Z., Leung, T., Li, L., Fei-Fei, L.: {{M}entor{N}et: Learning Data-Driven Curriculum for Very Deep Neural Networks on Corrupted Labels}. In: {International Conference on Machine Learning (ICML)} (2018)

\bibitem{2020_ICML_MentorMix}
Jiang, L., Huang, D., Liu, M., Yang, W.: {Beyond Synthetic Noise: Deep Learning on Controlled Noisy Labels}. In: {International Conference on Machine Learning (ICML)} (2020)

\bibitem{2016_CVPR_ResNet}
Kaiming, H., Xiangyu, Z., Shaoqing, R., Jian, S.: {Deep Residual Learning for Image Recognition}. In: {IEEE Conference on Computer Vision and Pattern Recognition (CVPR)} (2016)

\bibitem{2024_arxiv_Fly}
Kim, H., Chang, H.S., Cho, K., Lee, J., Han, B.: {Learning with Noisy Labels: Interconnection of Two Expectation-Maximizations}. arXiv: 2401.04390  (2024)

\bibitem{2009_CIFAR}
Krizhevsky, A., Hinton, G.: {Learning multiple layers of features from tiny images}. Tech. rep., University of Toronto (2009)

\bibitem{2012_NeurIPS_ImageNet}
Krizhevsky, A., Sutskever, I., Hinton, G.: {Imagenet classification with deep convolutional neural networks}. In: {Advances in neural information processing systems (NeurIPS)} (2012)

\bibitem{2021_ICLR_iMix}
Lee, K., Zhu, Y., Sohn, K., Li, C.L., Shin, J., Lee, H.: {i-Mix: A Strategy for Regularizing Contrastive Representation Learning}. In: {International Conference on Learning Representations (ICLR)} (2021)

\bibitem{2020_ICLR_DivideMix}
Li, J., Socher, R., Hoi, S.: {DivideMix: Learning with Noisy Labels as Semi-supervised Learning}. In: {International Conference on Learning Representations (ICLR)} (2020)

\bibitem{2021_ICCV_RRL}
Li, J., Xiong, C., Hoi, S.C.: {Learning from noisy data with robust representation learning}. In: {IEEE/CVF International Conference on Computer Vision (ICCV)} (2021)

\bibitem{2017_arXiv_WebVision}
{Li}, W., {Wang}, L., {Li}, W., {Agustsson}, E., {Van Gool}, L.: {WebVision Database: Visual Learning and Understanding from Web Data}. arXiv: 1708.02862  (2017)

\bibitem{2020_NeurIPS_EarlyReg}
Liu, S., Niles-Weed, J., Razavian, N., Fernandez-Granda, C.: {Early-Learning Regularization Prevents Memorization of Noisy Labels}. In: {Advances in Neural Information Processing Systems (NeurIPS)} (2020)

\bibitem{2021_CVPR_MOIT}
Ortego, D., Arazo, E., Albert, P., O'Connor, N.E., McGuinness, K.: {Multi-Objective Interpolation Training for Robustness to Label Noise}. In: {IEEE Conference on Computer Vision and Pattern Recognition (CVPR)} (2021)

\bibitem{2021_ICPR_towardsrobust}
Ortego, D., Arazo, E., Albert, P., O'Connor, N.E., McGuinness, K.: {Towards robust learning with different label noise distributions}. In: {International Conference on Pattern Recognition (ICPR)} (2021)

\bibitem{2021_ICML_CLIP}
Radford, A., Kim, J.W., Hallacy, C., Ramesh, A., Goh, G., Agarwal, S., Sastry, G., Askell, A., Mishkin, P., Clark, J., et~al.: {Learning transferable visual models from natural language supervision}. In: {International Conference on Machine Learning (ICML)} (2021)

\bibitem{2020_WACV_EDM}
Sachdeva, R., Cordeiro, F.R., Belagiannis, V., Reid, I., Carneiro, G.: {EvidentialMix: Learning with Combined Open-set and Closed-set Noisy Labels}. In: {IEEE/CVF Winter Conference on Applications of Computer Vision (WACV)} (2020)

\bibitem{2023_PR_scanmix}
Sachdeva, R., Cordeiro, F.R., Belagiannis, V., Reid, I., Carneiro, G.: {ScanMix: learning from severe label noise via semantic clustering and semi-supervised learning}. {Pattern Recognition}  (2023)

\bibitem{2020_arXiv_FixMatch}
Sohn, K., Berthelot, D., L, C.L., Zhang, Z., Carlini, N., Cubuk, E., Kurakin, A., Zhang, H., Raffel, C.: {FixMatch: Simplifying Semi-Supervised Learning with Consistency and Confidence}. arXiv: 2001.07685  (2020)

\bibitem{2019_ICML_SELFIE}
Song, H., Kim, M., Lee, J.G.: {{SELFIE}: Refurbishing Unclean Samples for Robust Deep Learning}. In: {International Conference on Machine Learning (ICML)} (2019)

\bibitem{2021_ICCV_weblyfinegrained}
Sun, Z., Yao, Y., Wei, X.S., Zhang, Y., Shen, F., Wu, J., Zhang, J., Shen, H.T.: {Webly Supervised Fine-Grained Recognition: Benchmark Datasets and An Approach}. In: {IEEE/CVF International Conference on Computer Vision (ICCV)} (2021)

\bibitem{2016_AAAI_Inception}
Szegedy, C., Ioffe, S., Vanhoucke, V., Alemi, A.: {Inception-v4, inception-resnet and the impact of residual connections on learning}. In: {Association for the Advancement of Artificial Intelligence (AAAI)} (2016)

\bibitem{2019_ICLR_Forgetting}
Toneva, M., Sordoni, A., Combes, R., Trischler, A., Bengio, Y., Gordon, G.: {An empirical study of example forgetting during deep neural network learning}. In: {International Conference on Learning Representations (ICLR)} (2019)

\bibitem{2022_JMLR_sololearn}
{Victor Guilherme Turrisi da Costa and Enrico Fini and Moin Nabi and Nicu Sebe and Elisa Ricci}: solo-learn: A library of self-supervised methods for visual representation learning. {Journal of Machine Learning Research}  (2022)

\bibitem{2016_NIPS_MiniImageNet}
Vinyals, O., Blundell, C., Lillicrap, T., Kavukcuoglu, K., Wierstra, D.: {Matching Networks for One Shot Learning}. In: {Advances in Neural Information Processing Systems (NeuRIPS)} (2016)

\bibitem{2023_IJCAI_promix}
Wang, H., Xiao, R., Dong, Y., Feng, L., Zhao, J.: {Promix: Combating label noise via maximizing clean sample utility}. In: International Joint Conference on Artificial Intelligence (IJCAI) (2023)

\bibitem{2020_ICLR_understandingcontrastive}
Wang, T., Isola, P.: {Understanding contrastive representation learning through alignment and uniformity on the hypersphere}. In: {International Conference on Machine Learning (ICLR)} (2020)

\bibitem{2022_ICLR_CIFARN}
Wei, J., Zhu, Z., Cheng, H., Liu, T., Niu, G., Liu, Y.: {Learning with Noisy Labels Revisited: A Study Using Real-World Human Annotations}. In: {International Conference on Learning Representations (ICLR)} (2023)

\bibitem{2021_CVPR_FaMUS}
Xu, Y., Zhu, L., Jiang, L., Yang, Y.: {Faster meta update strategy for noise-robust deep learning}. In: {IEEE/CVF Conference on Computer Vision and Pattern Recognition (CVPR)} (2021)

\bibitem{2021_CVPR_JoSRC}
Yao, Y., Sun, Z., Zhang, C., Shen, F., Wu, Q., Zhang, J., Tang, Z.: {Jo-SRC: A Contrastive Approach for Combating Noisy Labels}. In: {IEEE/CVF Conference on Computer Vision and Pattern Recognition (CVPR)} (2021)

\bibitem{2019_CVPR_PENCIL}
Yi, K., Wu, J.: {Probabilistic End-to-end Noise Correction for Learning with Noisy Labels}. In: {IEEE Conference on Computer Vision and Pattern Recognition (CVPR)} (2019)

\bibitem{2021_NeurIPS_flexmatch}
Zhang, B., Wang, Y., Hou, W., Wu, H., Wang, J., Okumura, M., Shinozaki, T.: {Flexmatch: Boosting semi-supervised learning with curriculum pseudo labeling}. In: {Advances in Neural Information Processing Systems (NeurIPS)} (2021)

\bibitem{2018_ICLR_Mixup}
Zhang, H., Cisse, M., Dauphin, Y., Lopez-Paz, D.: {mixup: Beyond Empirical Risk Minimization}. In: {International Conference on Learning Representations (ICLR)} (2018)

\bibitem{2021_ICLR_featurenoise}
Zhang, Y., Zheng, S., Wu, P., Goswami, M., Chen, C.: {Learning with feature-dependent label noise: A progressive approach}. In: {International Conference on Learning Representations (ICLR)} (2021)

\bibitem{2023_ICCV_rankmatch}
Zhang, Z., Chen, W., Fang, C., Li, Z., Chen, L., Lin, L., Li, G.: {RankMatch: Fostering Confidence and Consistency in Learning with Noisy Labels}. In: {IEEE/CVF International Conference on Computer Vision (ICCV)} (2023)

\bibitem{2022_WACV_C2D}
Zheltonozhskii, E., Baskin, C., Mendelson, A., Bronstein, A.M., Litany, O.: {Contrast to divide: Self-supervised pre-training for learning with noisy labels}. In: {IEEE/CVF Winter Conference on Applications of Computer Vision (WACV)} (2022)

\end{thebibliography}
